%% file: main.tex
\documentclass{article}
\usepackage{ijcai19}

\usepackage{times}
\usepackage{soul}
\usepackage{url}
\usepackage[hidelinks]{hyperref}
\usepackage[utf8]{inputenc}
\usepackage[small]{caption}
\usepackage{graphicx}
\usepackage{amsmath}
\usepackage{booktabs}
\usepackage{algorithm}
\usepackage{algorithmic}
\usepackage[normalem]{ulem}
\usepackage{color}
\newcommand{\pw}[1]{#1}
\usepackage{amsfonts}%mathbb
\usepackage{amsthm}%proof
\newtheorem{theorem}{Theorem}[section]
\newtheorem{lemma}[theorem]{Lemma}

\makeatletter
\newcommand{\pushright}[1]{\ifmeasuring@#1\else\omit\hfill$\displaystyle#1$\fi\ignorespaces}
\newcommand{\pushleft}[1]{\ifmeasuring@#1\else\omit$\displaystyle#1$\hfill\fi\ignorespaces}
\makeatother
\title{Exploiting the Sign of the Advantage Function to\\
Learn Deterministic Policies in Continuous Domains}

\author{
Matthieu Zimmer%$^1$
\And
Paul Weng%$^1$
\affiliations
UM-SJTU Joint Institute\\
\emails
\{matthieu.zimmer, paul.weng\}@sjtu.edu.cn
}

\begin{document}

\maketitle

\begin{abstract}
In the context of learning deterministic policies in continuous domains, we revisit an approach, which was first proposed in Continuous Actor Critic Learning Automaton (CACLA) and later extended in Neural Fitted Actor Critic (NFAC).
This approach is based on a policy update different from that of deterministic policy gradient (DPG).
Previous work has observed its excellent performance empirically, but a theoretical justification is lacking.
To fill this gap, we provide a theoretical explanation to motivate this unorthodox policy update by relating it to another update and making explicit the objective function of the latter.
We furthermore discuss in depth the properties of these updates to get a deeper understanding of the overall approach.
In addition, we extend it and propose a new trust region algorithm, Penalized NFAC (PeNFAC).
Finally, we experimentally demonstrate in several classic control problems that it surpasses the state-of-the-art algorithms to learn deterministic policies.
\end{abstract}

\section{Introduction}

\input{tex/intro.tex}

\section{Background}

\input{tex/background.tex}

\section{Algorithms}

\input{tex/algo.tex}

\section{Discussions}

\input{tex/discussion.tex}

\section{Extension to Trust Region}

\input{tex/penfac.tex}

\section{Experiments}

\input{tex/exp.tex}

\section{Conclusion}

In the context of learning deterministic policies, we studied the properties of two not very well-known but efficient updates, Continuous Actor Critic Learning Automaton (CACLA) and Continuous Actor Critic (CAC).
We first showed how closely they both are related to the stochastic policy gradient (SPG).
We explained why they are well designed to learn continuous deterministic policies when the value function is only approximated. 
We also highlighted the limitations of those methods: a potential poor sample efficiency when the dimension of the action space increases and no guarantee that the underlying deterministic policy will converge toward a local optimum of $J(\mu_\theta)$ even with a linear approximation.

In the second part, we extended Neural Fitted Actor Critic (NFAC), itself an extension of CACLA, with a trust region constraint designed for deterministic policies and proposed a new algorithm, Penalized NFAC (PeNFAC). 
Finally, we tried our implementation on various high-dimensional continuous environments and showed that PeNFAC performs better than DDPG and PPO to learn continuous deterministic policies.

As future work, we plan to consider off-policy learning and the combination of the updates of CAC and DPG together to ensure the convergence toward a local optimum while benefiting from the good updates of CAC.

\section*{Acknowledgments}

This work has been supported in part by the program of National Natural Science Foundation of China (No. 61872238).
Experiments presented in this paper were carried out using the Grid’5000 testbed, supported by a scientific interest group hosted by Inria and including CNRS, RENATER and several Universities as well as other organizations (see https://www.grid5000.fr).

%% The file named.bst is a bibliography style file for BibTeX 0.99c
\bibliographystyle{named}
\bibliography{ijcai19,manually_added}

\cleardoublepage

%commented for IJCAI deposit
\appendix
\input{tex/appendix.tex}

\end{document}

%% file: tex/intro.tex
Model-free reinforcement learning combined with neural networks achieved several recent successes over a large range of domains 
%where interactive data can be easily obtained 
\cite{Mnih2015,Lillicrap2015,PPO}.
Yet those methods are still difficult to apply without any expert knowledge, lack robustness and are very sensitive to hyperparameter optimization \cite{Henderson2017,colas2018gep}.

In this context, we focus in this paper on improving methods that learn deterministic policies.
Such policies have three main advantages during the learning phase:
1) they usually require less interactive data because fewer parameters need to be learned,
2) their performances are less costly to estimate during testing phases because randomness only comes from the environment (as opposed to randomized policies), and 
3) they are also less sensitive to the premature convergence problem, because they cannot directly control exploration. 
Moreover, deterministic policies are preferred in some domains (e.g., robotics), because we do not want the agent to act stochastically after the learning phase.
%Learning a deterministic policy is appealing 

In continuous state and action space domains, solution methods require function approximation.
Neural control architectures are %well fitted to represent the policy 
excellent representations for policies because they can handle continuous domains, are easily scalable, and have a high degree of expressiveness. 
The weights of such neural networks are usually updated with a policy gradient method.
As vanilla policy gradient suffers from high variance,
it is generally implemented in an actor-critic architecture where an estimated value function helps to reduce the variance at the cost of introducing some bias \cite{Konda1999}. 
In this architecture, the  parameters (e.g., weights of neural networks) of the policy (i.e., actor) and its value function (i.e., critic) are updated simultaneously. 

%Once the representation has been chosen, actor-critic methods are usually used to \pw{learn} the associated parameters \pw{(i.e., weights of neural networks)} of the policy and the critic. 
The basic version of an actor-critic architecture for learning deterministic policies in continuous domains is the deterministic policy gradient (DPG) method % where the parameters of the policy are adapted according to the gradient of the performance of the policy
\cite{Silver2014}. 
%The advantage function can locally tell if an action would be better or not than the one taken by the current policy.
Learning the value function is crucial but also difficult, which is why several extensions of DPG have been proposed. 
%The one which is the most currently used is
Deep Deterministic Policy Gradient (DDPG) \cite{Lillicrap2015} brings batch normalization \cite{Ioffe2015}, target networks and replay buffer \cite{Mnih2015} to DPG and is one of the most used actor-critic methods for learning continuous deterministic policies. 
However, it has several limitations: 
1) the critic learns the state-action value function (Q function), which is difficult to estimate, 
2) it relies on the fact that non-biased estimates of the gradient of the Q function are accessible, which is not the case in the model-free setting, 
3) it does not use compatible functions: 
the policy gradient %of the performance 
might be poorly estimated.

In this work, we focus on an alternative method that estimates the state value function (V function) instead of the Q function to learn continuous deterministic policies.
\citeauthor{VanHasselt2007} \shortcite{VanHasselt2007} were the first to propose to reinforce the policy toward an action with a positive temporal difference. % in an incremental setting.
They experimentally showed that using such a method, in an incremental 
actor-critic algorithm, called Continuous Actor Critic Learning Automaton (CACLA), provided better results than both the stochastic and the deterministic policy gradients\footnote{In their paper the deterministic policy gradient algorithm was called ADHDP \cite{Prokhorov1997}.} in the \textit{Mountain Car} and the \textit{Acrobot} environments.
\citeauthor{zimmer2016} \shortcite{zimmer2016,zimmer2018developmental} validated those results in higher-dimensional environments, \textit{Half-Cheetah} and \textit{Humanoid} in Open Dynamic Engine \cite{Smith2005}, 
and proposed several extensions with the Neural Fitted Actor Critic (NFAC) algorithm. 
However, no theoretical explanation for their good performance, nor a clear discussion about which objective function those methods optimize were given. 
Providing such an explanation would help understand better
why those algorithms work well, 
what are their properties and limitations, and 
how to further improve them.

We first show that CACLA and NFAC can be viewed as policy gradient methods and that they 
%and CAC 
are closely related to a specific form of the stochastic policy gradient (SPG) \cite{Sutton1999}. 
Then we discuss some of their properties and limitations.
Moreover, we extend them with trust region updates and call the new algorithm %, and propose a new trust region extension \pw{of them, called} 
Penalized Neural Fitted Actor Critic (PeNFAC). 
Finally, we experimentally show that PeNFAC performs well on three high-dimensional continuous environments compared to the state-of-the-art methods.

%% file: tex/background.tex
A continuous Markov Decision Process (MDP) \cite{sutton1988learning} is a tuple $(\mathcal{S}, \mathcal{A}, T,R,T_0)$ where 
$\mathcal{S}$ is a continuous state space, 
$\mathcal{A}$ is a continuous action space with $m$ dimensions, 
$T:\mathcal S \times \mathcal A \times \mathcal S \to [0, 1]$ is a %Markovian stationary 
transition function, %dynamics distribution,
$R: \mathcal{S} \times \mathcal A \rightarrow \mathbb{R}$ is a reward function, % that maps state and action to a scalar,
$T_0$ is a distribution over initial states. % to define the first state.
In the model-free setting, it is assumed that the transition function $T$ and the reward function $R$ are unknown and can only be sampled at specific states according to the interaction between the agent and the environment.

The following notations are used: 
$\mu$ represents a deterministic policy and 
$\pi$ a stochastic one. 
Thus, for a given state $s \in \mathcal{S}$, $\mu(s)$ is an action, $\pi(a|s)$ is the probability of sampling action $a$ from the policy $\pi$, and $\pi(\cdot|s)$ is a distribution over the action space $\mathcal{A}$.
For a policy $\pi$, we denote the discounted state distribution by: 
\begin{align*}
    d^\pi_\gamma(s) = \int_{\mathcal{S}} T_0(s_0)\sum^\infty_{t=0}   \gamma^{t} p(s|s_0,t,\pi) ds_0
\end{align*} 
where $\gamma \in [0, 1)$ is a discount factor and $p(s|s_0,t,\pi)$ is the probability of being in state $s$ after applying policy $\pi$ $t$ timesteps from state $s_0$.
Its state value function is defined by
%\begin{equation*}
$V^\pi(s)$ $=$ $\mathbb{E}_\pi \big[ \sum_{t=0}^{\infty} \gamma^{t} R(S_t, A_t) \big| S_0 = s \big]$
where $\mathbb E_\pi$ is the expectation induced by $T$ and $\pi$, and for all $t$, $S_t$ and $A_t$ are random variables. 
Its action value function is given by
%\begin{equation*}
$Q^\pi(s,a) = \mathbb{E}_\pi \big[ \sum_{t=0}^{\infty} \gamma^{t} R(S_t, A_t) \big| S_0 = s, A_0 = a \big],$ 
and its advantage function by $A^\pi(s,a)=Q^\pi(s,a)-V^\pi(s)$.
%\end{equation*}

In reinforcement learning, the goal is to find a policy that optimizes the expectation of the discounted rewards: 
\begin{equation*}
J(\pi) = \mathbb{E}_\pi \Big[ \sum_{t=0}^\infty \gamma^t R(S_t, A_t) \big| S_0 \sim T_0 \Big].
\end{equation*}
Due to the continuity of the state/action spaces, %we consider the problem of optimizing 
this optimization problem is usually restricted to a class of parametrized policies, which we denote $\pi_{\theta}$ (stochastic case) or $\mu_\theta$ (deterministic case). 
To simplify notations, we may write $\pi$ or $\mu$ instead of $\pi_{\theta}$ or $\mu_\theta$. 
The stochastic policy gradient (SPG) in the continuous case can be written as \cite{Sutton1999}:
%Deterministic on-policy objective:
\begin{equation}
\label{eq:SPG}
\nabla_\theta J(\pi) = \int_{\mathcal{S}} d^\pi_\gamma(s) \int_{\mathcal{A}} A^\pi(s,a) \nabla_\theta \pi_\theta(a|s)  da ds.
\end{equation}
The DPG is defined as \cite{Silver2014}: 
\begin{align}
\nabla_\theta J(\mu) &= \int_{\mathcal{S}} d^\mu_\gamma(s) \Delta_{\text{DPG}}(s,\mu_\theta) ds, \\
\text{where} \notag\\
 \Delta_{\text{DPG}}(s,\mu_\theta) &= \nabla_a A^\mu(s,a) \big|_{a=\mu_\theta(s)} \nabla_\theta \mu_\theta(s).
\end{align}
Policy gradient methods usually take a step according to those directions: $\theta_{t+1} \leftarrow \theta_t + \alpha \nabla_\theta J$. 
However, it is difficult to select a proper learning rate $\alpha$ to control the step size. 
If $\alpha$ is too big, %the performance of the policy might be decreased. 
the method may diverge.
If it is too low, the learning will converge slowly (thus requiring more samples).
To overcome this difficulty, a trust region method can be used to control the step size \cite{Schulman2015}. %\cite{K
Indeed, one can guarantee monotonic gradient updates by exploiting an approximation of the policy advantage function \cite{kakade2002approximately} of $\tilde{\pi}$ with respect to $\pi$, which measures the difference of performance between the two policies:
% It easily extends to deterministic policies as follow:
%\begin{equation}
%J(\tilde{\mu}) = J(\mu) + \int_{\mathcal{S}} d^{\tilde{\mu}}_\gamma(s) A^\mu(s, \tilde{\mu}(s)) ds.
%\end{equation}
\begin{align}
\label{eq:stochperfmeasure}
J(\tilde{\pi}) =& J(\pi) + \int_{\mathcal{S}} d^{\tilde{\pi}}_\gamma(s) \int_{\mathcal{A}} \tilde{\pi}(a|s) A^\pi(s, a) da ds, \\
\approx & J(\pi) + \int_{\mathcal{S}} d^{\pi}_\gamma(s) \int_{\mathcal{A}} \tilde{\pi}(a|s) A^\pi(s, a) da ds. \notag
\end{align}
The latter approximation holds when the two policies are close, which can be enforced by a KL divergence constraint in  trust region policy optimization \cite{Schulman2015}.
%\begin{equation}
%J(\tilde{\mu}) = J(\mu) + \int_{\mathcal{S}} d^{\tilde{\mu}}_\gamma(s) A^\mu(s, \tilde{\mu}(s)) ds.
%\end{equation}
%
%Thus if we can ensure that the right part of the equation is positive, the new policy $\tilde{\mu}$ will be better than the current policy $\mu$.
%Repeating the process and it will converge toward a local minima.
%
%However, it is difficult to optimize the right part in an online reinforcement learning setting because the new discounted distribution $d^{\tilde{\mu}}_\gamma$ is unknown. One usually have only access to an approximation of the current discounted distribution. 
%
%To keep $\tilde{\pi}$ and $\pi$ stay relatively close, in order to guarantee that $\tilde{\pi}$ will be better than $\pi$, \cite{Schulman2015} proposed to constrained the Kullback–Leibler divergence of the two policies for each taken policy improvement step.

%% file: tex/algo.tex
In this section, we recall three related algorithms (CACLA, CAC, NFAC) that we discuss later.

\subsection{Continuous Actor Critic Learning Automaton}
%They also introduce Continuous Actor Critic (CAC)

Continuous Actor Critic Learning Automaton (CACLA) \cite{VanHasselt2007} is an actor-critic method that learns a stochastic policy $\pi$ and its estimated value function $\hat V^\pi$.
%A stochastic policy $\pi$ %built on the current deterministic one,
%is used to ensure \pw{sufficient} exploration during learning. % in the exploration-exploitation dilemma.
We assume in this paper that CACLA uses isotropic Gaussian exploration, which implies that %the exploratory stochastic policy takes the following form:
$\pi$ can be written as follows:
\begin{equation}
\label{eq:hypo_polstoch}
\pi_{\theta,\sigma}(\cdot|s) = \mathcal{N}\big(\mu_\theta(s), \sigma^2 I)
\end{equation}
where $I$ is the identity matrix and $\sigma>0$ possibly annealed during learning. %$\sigma>0$ is a fixed variance (possibly annealed during learning).
CACLA alternates between two phases:

\noindent 1) a hill climbing step in the action space using a random optimization (RO) algorithm \cite{matyas1965random}, 

\noindent 2) a gradient-like update in the policy parameter space.
RO consists in repeating the following two steps: 

i) sample a new action $a'$, which is executed in the environment in current state $s$, by adding a normally distributed noise to the current action $a=\mu(s)$, 

ii) if $R(s, a') + \gamma \hat V^\pi(s') > \hat V^\pi(s)$ then $a \leftarrow a'$ else $a$ does not change.

\noindent Phase 2) is based on following update: 
\begin{equation} \label{eq:base_cacla}
  \text{If } \delta(s,a) > 0: \tilde{\theta} \leftarrow \theta - \alpha \big(\mu_\theta(s) - a\big) \nabla_\theta \mu_\theta(s),  
\end{equation}
where $\delta(s,a) = R(s, a) + \gamma \hat V^\pi(s') - \hat V^\pi(s)$ is the temporal difference (TD) error. 
As the expectation of the TD error is equal to the advantage function, this update can be interpreted as follows: if an exploratory action $a$ has a positive advantage then policy $\mu$ should be updated towards $a$. 

Note that although CACLA executes a stochastic policy $\pi$, it can be seen as learning a deterministic policy $\mu$.
\citeauthor{VanHasselt2007} \shortcite{VanHasselt2007} state that when learning in continuous action space, moving away from a bad action could be meaningless. 
%It is especially true when the learned policy is deterministic because 
Indeed, while for stochastic policies, the probability of a bad action can be decreased, %(without moving toward an uncertain direction). 
for deterministic policies, moving in the action space in the opposite direction of an action with a negative advantage may not necessarily lead to better actions.
Thus, CACLA's update is particularly appropriate for learning continuous deterministic policies.

%\pw{Note that although a deterministic policy is learned, the} policy ultimately executed \pw{in} the environment during the learning phase is $\pi_{\theta,\sigma}$.
%Because the temporal difference is also used to update the state value function V, the critic which is learned is $V^\pi$ even though $\mu_\theta$ is updated.

\subsection{Continuous Actor Critic}

In our discussion, we also refer to a slightly different version of CACLA, Continuous Actor Critic (CAC) \cite{VanHasselt2007}. 
The only difference between CAC and CACLA is that 
%the scale of the advantage function is used. 
the update in CAC is scaled by the TD error:
\begin{equation}
  \text{If } \delta(s,a) > 0: \tilde{\theta} \leftarrow \theta - \alpha \delta(s,a) \big(\mu_\theta(s) - a\big) \nabla_\theta \mu_\theta(s),  
\end{equation}
Thus an action with a larger positive advantage (here, estimated by the TD error) will have a bigger impact over the global objective.

\subsection{Neural Fitted Actor Critic}

The Neural Fitted Actor Critic (NFAC)  \cite{zimmer2016,zimmer2018developmental} algorithm is an efficient instantiation of the CACLA update, which integrates the following techniques: batch normalization, $\lambda$-returns for both the critic and the actor, and batch learning with Adam\cite{Kingma2015}. 
In this algorithm, the update of the parameters is not done anymore at each time step, but at the end of a given number of episodes.

%% file: tex/discussion.tex
In this section, we discuss the algorithms to provide some theoretical explanation for their good performance.

\subsection{CACLA}

We first explain the relationship between an algorithm based on stochastic policy gradient (SPG) and CACLA. %(this is the way it works the best).
For this discussion, we assume that SPG is applied to parametrized policies that are Gaussian policies $\pi_{\theta, \sigma}$ (i.e., Gaussian around $\mu_\theta$). 
Then the first common feature between the two algorithms is that the distributions over states they induce during learning are the same (i.e., $d^{\pi}_\gamma(s)$) because they both use the same exploratory policy to interact with the environment.
Moreover, SPG can be written as follows:
\begin{align*}
&\nabla_\theta J(\pi_{\theta,\sigma}) \notag \\ 
&= \int_{\mathcal{S}} d^\pi_\gamma(s) \int_{\mathcal{A}} \pi_{\theta,\sigma}(a|s) A^\pi(s,a) \nabla_\theta \text{ log } \pi_\theta(a|s)  da ds, \notag  \\
&= \frac{1}{\sigma^2} \int_{\mathcal{S}} d^\pi_\gamma(s) \int_{\mathcal{A}} \pi_{\theta,\sigma}(a|s) A^\pi(s,a) \big(a - \mu_\theta(s)\big) \cdot \\ 
& \ \ \ \ \ \ \ \ \ \ \ \ \ \ \ \ \ \ \ \ \ \ \ \ \ \ \ \ \ \ \ \ \ \ \ \ \ \ \ \ \ \ \ \ \ \ \ \ \ \ \ \ \ \ \ \  \ \ \ \ \ \ \ \ \ \ \ \ \ \  \nabla_\theta \mu_\theta(s)  da ds \notag.
\end{align*}
For CACLA, we interpret update~(\ref{eq:base_cacla}) as a stochastic update in the following direction:
\begin{flalign}
\label{eq:caclaeq}
& \int_{\mathcal{S}} d^{\pi}_\gamma(s) \Delta_{\text{CACLA}}(s, \mu_\theta) ds,\\ \text{with } & \Delta_{\text{CACLA}}(s,\mu_\theta) = 
\int_{\mathcal{A}} \pi_{\theta, \sigma}(a|s) 
H\big(A^\pi(s,a)\big) \times \notag \\
& \hspace{12em}
\big(\mu_\theta(s) - a\big) \nabla_\theta \mu_\theta(s) da \notag,
\end{flalign}
% \begin{flalign}
% \label{eq:caclaeq}
% %J^{\text{CACLA}}(\mu) =
%  \Delta_{\text{CACLA}}(\mu_\theta) & =
%  \int_{\mathcal{S}} d^{\pi}_\gamma(s)
%  \int_{\mathcal{A}} \pi_{\theta,\sigma}(a|s) \notag \\
% &  H(A^\pi(s,a)) (\mu_\theta(s) - a) \nabla_\theta \mu_\theta(s) da ds,
% \end{flalign}
where $H$ is the Heaviside function. 
Indeed, the inner integral is estimated using a single Monte Carlo sample during the run of CACLA. % in Equation \ref{eq:base_cacla}. 

Under this form, it is easy to see the similarity between SPG and CACLA. % are near using (\ref{eq:SPG}) with (\ref{eq:hypo_polstoch}): 
%
%For the critic part, notice that the temporal difference error:
%Rewriting the temporal difference error as
%\begin{align*}
%\delta(s,a) &= \mathbb{E}_{s' \sim T(\cdot|s,a), a \sim %\pi(\cdot|s)}[R(s) + \gamma V(s') ] - V(s),\\
%&= Q^\pi(s,a) - V^\pi(s),\\
%&= A^\pi(s,a),
%\end{align*}
%leads to denote the CACLA update direction as:
%
The constant factor $\frac{1}{\sigma^2}$ can be neglected because it may be integrated into the learning rate. 
The sign difference of the term $(a-\mu_\theta(s))$ is because SPG performs gradient ascent and CACLA  gradient descent. 
So the main difference between SPG and CACLA is the replacement of $A^\pi(s,a)$ by $H(A^\pi(s,a))$. 
Therefore CACLA optimizes its exploratory stochastic policy through an approximation of SPG hoping to improve the underlying deterministic policy (for a fixed state, the direction of CACLA and SPG are the same up to a scalar).

Moreover, relating CACLA's update with (\ref{eq:caclaeq}) also brings to light two main limitations. 
The first one concerns the inner integral over the action space which has a high variance. 
Therefore, we expect CACLA to be less and less data efficient in high-dimension action space (which is the main theoretical justification of DPG over SPG - see Appendix~\ref{appendix:sensitivitydima}). 
The second limitation that appears is that over one update, CACLA does not share the same exact optimal solutions as DPG or SPG. 
%However, in the general case, this cost doesn't share the exact same solutions of $L_\mu(\tilde{\mu})$ neither of $J(\mu)$.
%In specific cases, where the expressive power of $\mu$ could reduce all $A^+(\pi,s)$ to zero, then the solutions will be shared.
Indeed, if we define $\theta^*$ such as $\nabla_\theta J(\mu_{\theta})\big|_{\theta =\theta^*} = 0$ it is not possible to prove that (\ref{eq:caclaeq}) will also be 0 (because of the integral over the state space). 
It means that CACLA could decrease the performance of this local optimal solution.

\subsection{CAC}

Similarly, the update in CAC can be seen as a stochastic update in the following direction:
\begin{flalign}
\label{eq:cac}
%\Delta_{\text{CAC}}(\mu_\theta) &=
& \int_{\mathcal{S}} d^{\pi}_\gamma(s) \Delta_{\text{CAC}}(s, \mu_\theta) ds, \notag \\ %\hspace{.7cm}
\text{with } & \Delta_{\text{CAC}}(s,\mu_\theta) = 
\int_{\mathcal{A}} \pi_{\theta, \sigma}(a|s) A^\pi(s,a) H\big(A^\pi(s,a)\big) \times \notag \\ & \hspace{10em} \big(\mu_\theta(s) - a\big) \nabla_\theta \mu_\theta(s) da \notag.
\end{flalign}
%where $M^\pi(s,a)$ is defined as $A^\pi(s,a) H(A^\pi(s,a))$.
This shows that CAC is even closer to SPG than CACLA and provides a good theoretical justification of this update at a local level (not moving in potentially worst action). 
However, there is also a justification at a more global level.
\begin{lemma} For a fixed state, when the exploration tends to zero, %the sign of the direction by CAC is the same as the one of DPG and the scale may only be smaller: 
CAC maintains the sign of the DPG update with a scaled magnitude:
\begin{equation}
\lim_{\sigma \rightarrow 0} \Delta_{\text{CAC}} (s, \mu_\theta) \gets g^+(s,\pi) \circ \Delta_{\text{DPG}} (s, \mu_\theta),
%\int_{\mathcal{A}} \pi(a|s)  (\mu_\theta(s) - a) A^\pi(s,a) H(A^\pi(s,a)) 
%\nabla_\theta \mu_\theta(s) da
\end{equation}
where $g^+(s,\pi)$ is a positive function between $[0; 1]^{n}$ with $n$ as the number of parameters of the deterministic policy and $\circ$ is the Hadamard product (element-wise product).
\end{lemma}
The proof is provided in Appendix~\ref{appendix:prooflimCAC}.
The consequence of this lemma is that, for a given state and low exploration, a local optimal solution for DPG will also be one for CAC. 
However it is still not the case for the overall update because of the integral over the different states. 
The weights given to each direction over different states are not the same in CAC and DPG.
One might think that in such a case, it would be better to use DPG.
However, in practice, the CAC update may in fact be more accurate when using an approximate advantage function. 
Indeed, there exist cases where DPG with an approximate critic  might update towards a direction which could decrease the performance.
%To understand why, the definition of $g^+(s,\pi)$ must be analyzed. 
%It will be equal to zero in state $s$ if all the actions in a neighborhood of $\mu(s)$ have a negative advantage. 
%In such a case, there are two possibilities regarding 
For instance, when the estimated advantage $\hat{A}\big(s,\mu(s) \big)$ is negative,
%\begin{itemize}
%    \item $\hat{A}(s,\mu(s)) = 0$: $\mu(s)$ is estimated as the optimal action in state $s$.
%    Therefore, both DPG and CAC will not update it, or 
%    \item $\hat{A}(s,\mu(s)) < 0$: 
the advantage around $\mu(s)$ is therefore known to be poorly estimated.
In such a case, thanks to the Heaviside function, 
CAC will not perform any update for actions $a$ in the neighborhood of $\mu(s)$ such that $\hat A(s, a) \le 0$. %because it is aware that the critic is poorly estimated at this point. 
However, in such a case, DPG will still perform an update according to this poorly estimated gradient. 
%\end{itemize}
%An example is given in Figure ...

%% file: tex/penfac.tex
In this section, we extend the approach to use a trust region method.

\subsection{Trust Region for Deterministic Policies}

We now introduce a trust region method dedicated to continuous deterministic policies.
Given current deterministic policy $\mu$, and an exploratory policy $\pi$ defined from $\mu$, the question is to find a new deterministic policy $\tilde{\mu}$ that improves upon $\mu$. 
%Similarly to \cite{Schulman2015}, we can introduce first-order surrogate objectives to solve this problem. However in the deterministic case, we often don't even have a direct access to the current discounted distribution $d^{\mu}_\gamma$ but to a stochastic exploratory version because of the exploration-exploitation dilemma.
%Let us write an exploratory stochastic policy from $\mu_\theta(s)$ as $\pi_{\theta(s),\sigma}(a|s)$ where $\pi$ as the following properties $\mathbb{E}[\pi(a|s)] = \mu(s)$...
%
Because a deterministic policy is usually never played in the environment outside of testing phases, a direct measure between two deterministic policies (i.e., a deterministic equivalent of Equation $\ref{eq:stochperfmeasure}$) is not directly exploitable. 
Instead we introduce the following measure:
\begin{lemma} 
       The performance $J(\tilde{\mu})$ of a deterministic policy $\tilde{\mu}$ can be expressed by the advantage function of another stochastic policy $\pi$ built upon a deterministic policy $\mu$ as:
       \begin{flalign} \label{eq:j mu bar}
       J(\tilde{\mu}) = J(\mu) + \int_{\mathcal{S}} d_\gamma^{\pi}(s) \int_\mathcal{A} \pi(a|s) A^\mu(s, a) da ds + \notag \\ 
       \int_{\mathcal{S}} d_\gamma^{\tilde{\mu}}(s) A^\pi \big(s, \tilde{\mu}(s)\big) ds.
       \end{flalign}
\end{lemma}
See Appendix~\ref{appendix:fosl} for the proof.
The first two quantities in the RHS of (\ref{eq:j mu bar}) are independent of ${\tilde{\mu}}$. 
The second one represents the difference of performance from moving from the deterministic policy $\mu$ to its stochastic version $\pi$. 
Because $d^{\tilde{\mu}}_\gamma$ would be too costly to estimate, we approximate it with the simpler quantity $d_\gamma^\pi$, as done by \citeauthor{Schulman2015} \shortcite{Schulman2015} for TRPO, a predecessor to PPO.
\begin{theorem} \label{theo:trustdeter} Given two deterministic policies $\mu$ and $\tilde{\mu}$, a stochastic Gaussian policy $\pi$ with mean $\mu(s)$ in state $s$ and independent variance $\sigma$, if the transition function $T$ is L-Lipschitz continuous with respect to the action from any state then:
\begin{flalign*}
    &\Big| \int_{\mathcal{S}} d^{\tilde{\mu}}(s) A^\pi \big(s, \tilde{\mu}(s)\big) - \int_{\mathcal{S}} d^{\pi}(s) A^\pi \big(s, \tilde{\mu}(s)\big)  \Big| \leq \\ & \frac{\epsilon L}{1-\gamma} \underset{t>0}{\operatorname{max\ }} \Big( \big|\big| \tilde{\mu}(s) - \mu(s)\big|\big|_{2,\infty} + \frac{2m\sigma}{\sqrt{2 \pi}} \Big)^t,
\end{flalign*}
where $\epsilon = \text{max}_{s,a} |A^\pi(s,a)| $.
\end{theorem}
\noindent The proof is available in Appendix~\ref{appendix:prooftrust}.
Thus, to ensure a stable improvement at each update, we need to keep both $|| \mu - \tilde{\mu} ||_{2,\infty}$ and $\sigma$ small. %By using G
Note that the Lipschitz continuity condition is natural in continuous action spaces.
It simply states that for a given state, actions that are close will produce similar transitions. % but the Lipschitz constraint might not hold in every state or in specific environment.

\subsection{Practical Algorithm}

%\paragraph{NFAC}
%NFAC uses the CACLA update and brings batch normalization, eligibility traces for both the critic and the actor and batch learning with ADAM.

To obtain a concrete and efficient algorithm,  the trust region method can be combined with the previous algorithms. 
Its integration to NFAC with a CAC update for the actor is called Penalized Neural Fitted Actor Critic (PeNFAC).

\citeauthor{VanHasselt2007} \shortcite{VanHasselt2007} observed that the CAC update performs worse that the CACLA update in their algorithms.
In their setting where the policy and the critic are updated at each timestep, we believe this observation is explained by the use of the TD error (computed from a single sample) to estimate the advantage function.
However, when using variance reduction techniques such as $\lambda$-returns and learning from a batch of interactions, or when
mitigating the update with a trust region constraint, we observe that this estimation becomes better (see~Figure~\ref{fig:penfaccomp}).
This explains why we choose a CAC update in PeNFAC.

In order to ensure that $|| \mu - \tilde{\mu} ||_{2,\infty}$ stays small over the whole state space, we approximate it with a Euclidean norm over the state visited by $\pi$. 
To implement this constraint, we add a regularization term to the update and automatically adapts its coefficient, for a trajectory $(s_0, s_1, \ldots, s_h)$:
\begin{equation*}
    %\Delta_{\text{PeNFAC}}(s_t, \mu_\theta) =
     \sum_{t=0}^{h-1} \Delta_{\text{CAC}}(s_t, \mu_\theta) + \beta \nabla_\theta \big|\big| \mu_{\text{old}}(s_t) - \mu_\theta(s_t) \big|\big|^2_2,
\end{equation*}
where $\beta$ is a regularization coefficient.
Similarly to the adaptive version of Proximal Policy Optimization (PPO) \cite{PPO}, $\beta$ is updated in the following way (starting from $\beta \leftarrow 1$):
\begin{itemize}
    \item if $\hat{d}(\mu,\mu_{\text{old}}) < d_{\text{target}} / 1.5$: $\beta \leftarrow \beta / 2 $,
    \item if $\hat{d}(\mu,\mu_{\text{old}}) > d_{\text{target}} \times 1.5$: $\beta \leftarrow \beta \times 2 $,
%    \item else stop,
\end{itemize}
where $\hat{d}(\mu,\mu_{\text{old}}) = \frac{1}{\sqrt{m L}} \sum_{s \sim \pi} || \mu_{\text{old}}(s) - \mu_\theta(s) ||_2$ with $L$ being the number of gathered states.
Those hyper-parameters are usually not optimized because the learning is not too sensitive to them.
The essential value to adapt for the designer is $d_\text{target}$.
Note that the introduction of this hyperparameter mitigates the need to optimize the learning rate for the update of the policy, which is generally a much harder task.

%% file: tex/exp.tex
We performed two sets of experiments to answer the following questions:

1) How does PeNFAC compare with state-of-the-art algorithms for learning deterministic policies?

2) Which components of PeNFAC contribute the most to its performance?

%first we analyze the performance of our proposed algorithm, then we check which component of PeNFAC is more essential to the algorithm.

The experiments were performed on environments with continuous state and action spaces in a time-discretized simulation.
%We \pw{chose} to benchmark state-of-the-art algorithms on
We chose to perform the experiments on
OpenAI Roboschool \cite{PPO},
a free open-source software, which allows anyone to easily reproduce our experiments.
%because it is a free and open-source environment where it is easy for anyone to reproduce experimentation.
In order to evaluate the performance of an algorithm, deterministic policies $\mu$ obtained during learning are evaluated at a constant interval during testing phases:
%In order to evaluate deterministic \pw{policies} obtained during the learning phase, at a constant interval, 
%during the testing phase, 
policy $\mu$ is played in the environment without exploration. 
The interactions gathered during this evaluation are not available to any algorithms. 
The source code of the PeNFAC algorithm is available at \url{github.com/matthieu637/ddrl}. 
The hyperparameters used are reported in Appendix \ref{appendix:hyperparam} as well as the considered range during the grid search.

\subsection{Performance of PeNFAC}

We compared the performance of PeNFAC to learn continuous deterministic policies with two state-of-the-art algorithms: PPO and DDPG. 
A comparison with NFAC is available in the ablation study (Section \ref{sec:ablationstudy}) and in Appendix \ref{appendix:penfacvsnfac}.
Because PPO learns a stochastic policy, for the testing phases, we built a deterministic policy as follows $\mu(s) = \mathbb{E}[a | a \sim \pi_\theta(\cdot,s)]$. 
We denote this algorithm as "deterministic PPO". 
In Appendix \ref{appendix:deterppo}, we experimentally show that %it does not hurt the performances of PPO.
this does not penalize the comparison with PPO, as deterministic PPO provides better results than standard PPO.
For PPO, we used the OpenAI Baseline implementation. To implement PeNFAC and compare it with NFAC, we use the DDRL library \cite{zimmer2018developmental}. Given that DDPG is present in those two libraries, we provided the two performances for it. 
The OpenAI Baseline version uses an exploration in the parameter space and the DDRL version uses n-step returns.

\begin{figure}[tb]
    \begin{center}
        \includegraphics[width=0.92\linewidth]{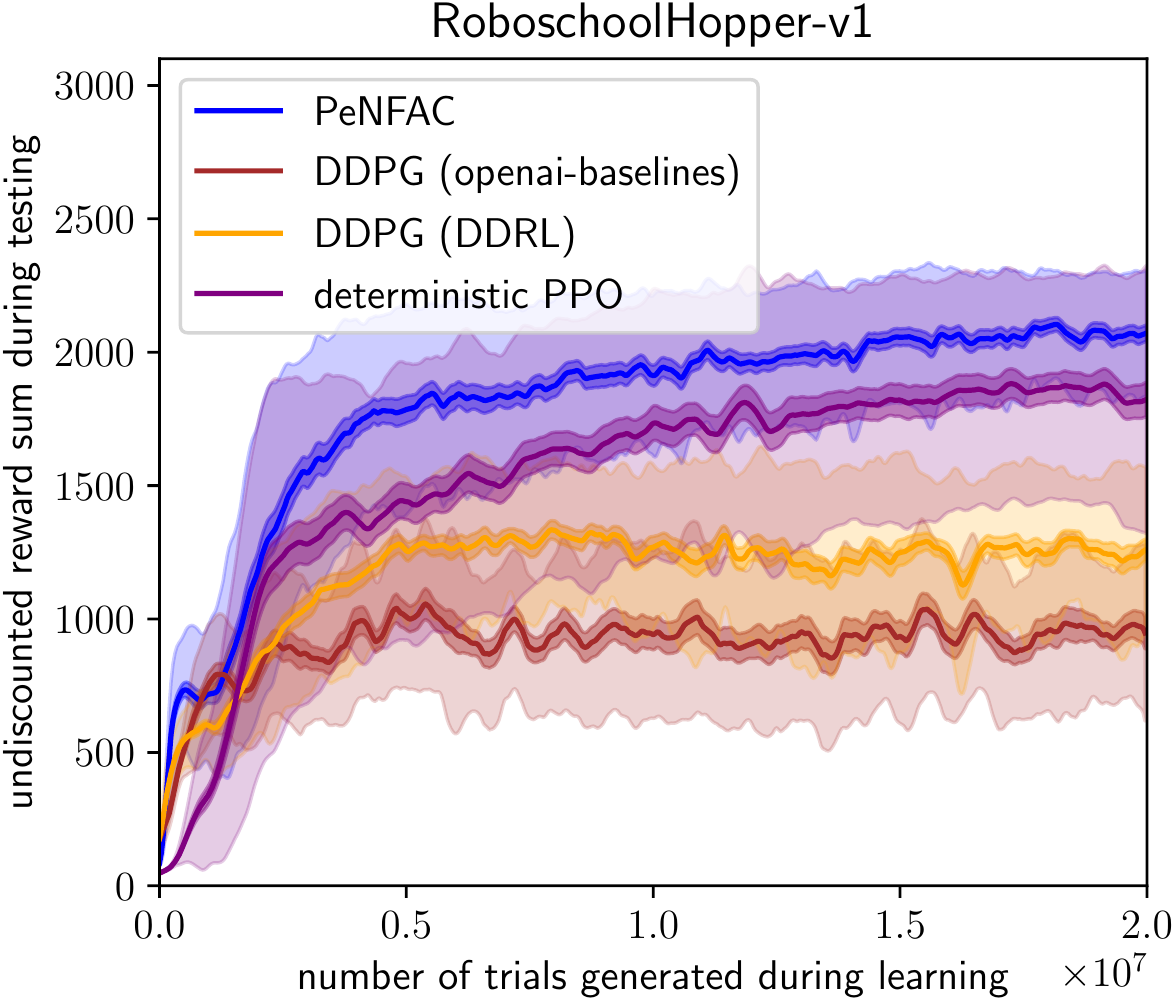}
    \end{center}
    \caption{Comparison of PeNFAC, DDPG and deterministic PPO over 60 different seeds for each algorithm in Hopper.}  
    \label{fig:penfacperf1}
\end{figure}
\begin{figure}[tb]
    \begin{center}
        \includegraphics[width=0.92\linewidth]{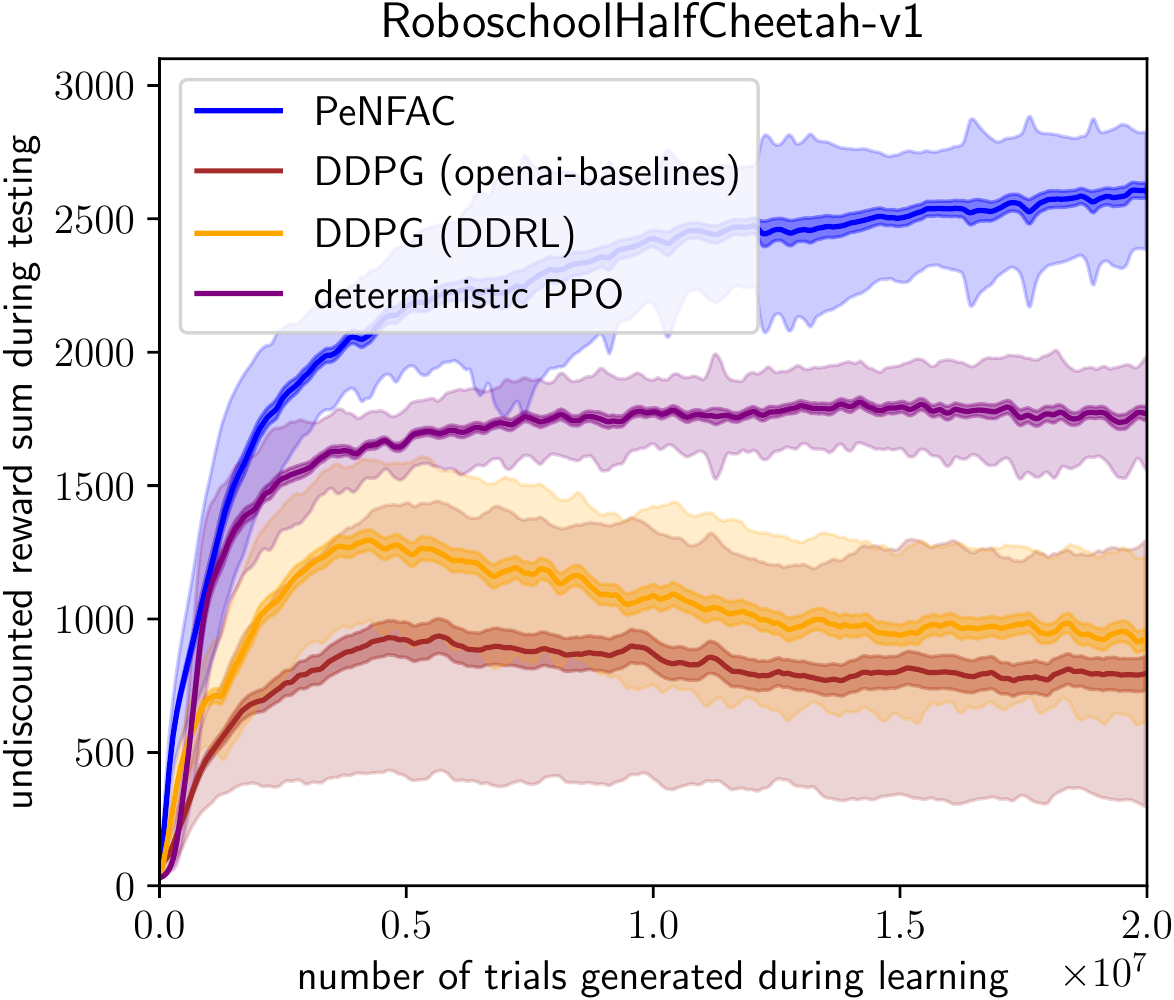}
    \end{center}
    \caption{Comparison of PeNFAC, DDPG and deterministic PPO over 60 different seeds for each algorithm in HalfCheetah.}  
    \label{fig:penfacperf2}
\end{figure}
\begin{figure}[H]
    \begin{center}
        \includegraphics[width=.92\linewidth]{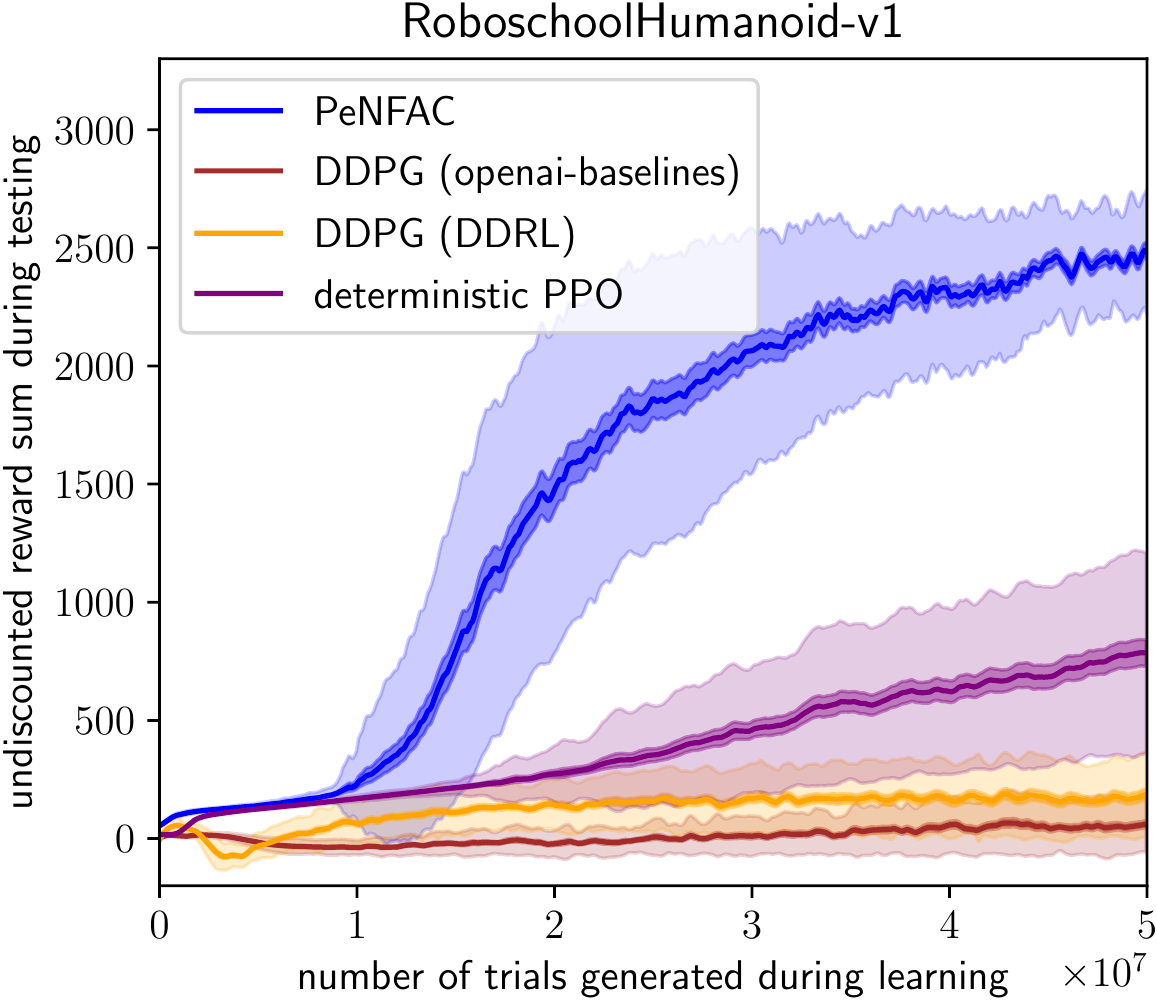}
    \end{center}
    \caption{Comparison of PeNFAC, DDPG and deterministic PPO over 60 seeds for each algorithm in Humanoid.}  
    \label{fig:penfacperf3}
\end{figure}

We performed learning experiments over three high-dimensional domains: 
Hopper, HalfCheetah and Humanoid.
Dimensions of $\mathcal{S} \times \mathcal{A}$ are $15 \times 3$ (Hopper),  $26 \times 6$ (HalfCheetah) and $44 \times 17$ (Humanoid).

The neural network architecture is composed of two hidden layers of 64 units for either the policy or the value function. 
The choice of the activation function in the hidden units was optimized for each algorithm: we found that ReLU was better for all of them except for PPO (where tanh was better). 
The output activation of the critic is linear and the output activation of the actor is tanh.

\begin{figure}[H]
    \begin{center}
        \includegraphics[width=.89\linewidth]{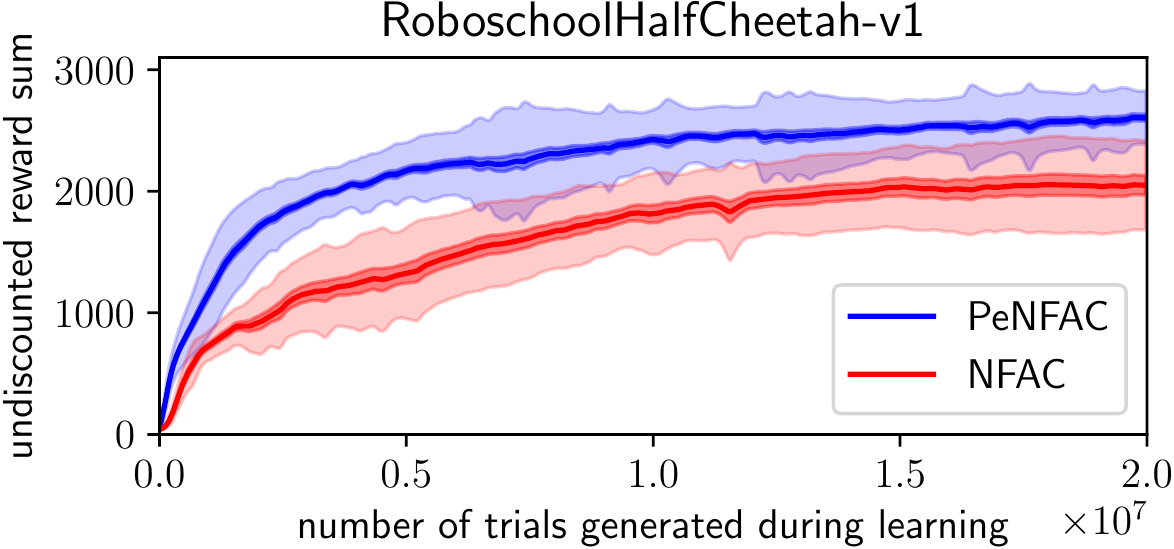}
        \includegraphics[width=.89\linewidth]{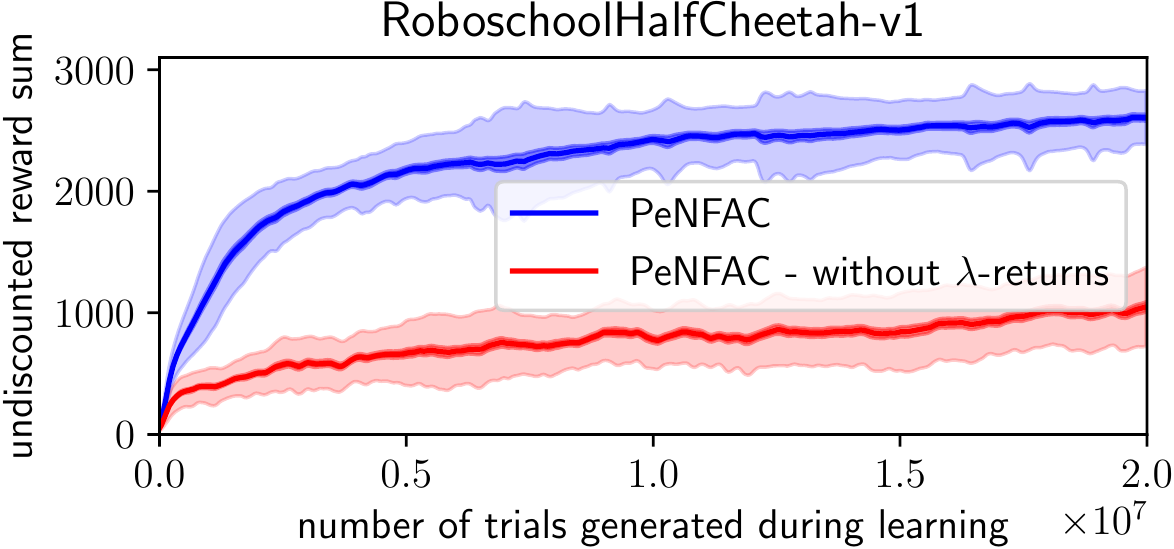}
        \includegraphics[width=.89\linewidth]{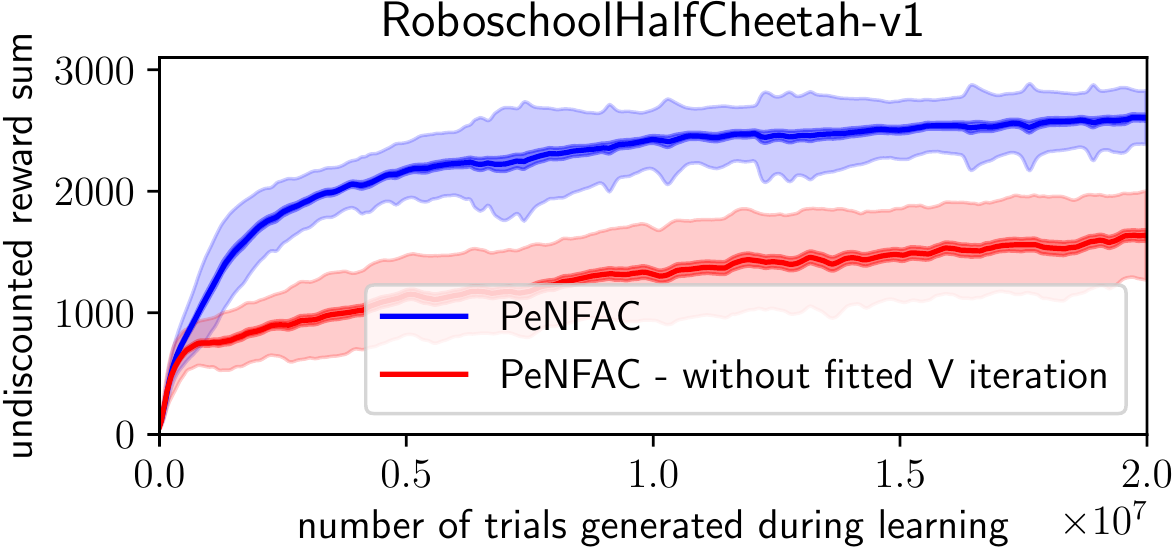}
        \includegraphics[width=.89\linewidth]{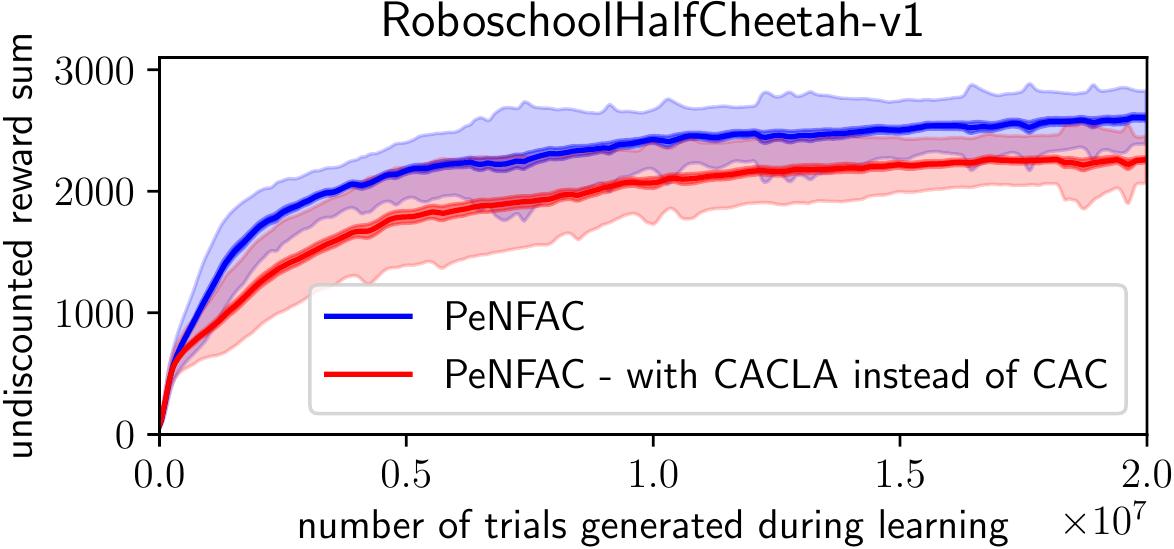}
        \includegraphics[width=.89\linewidth]{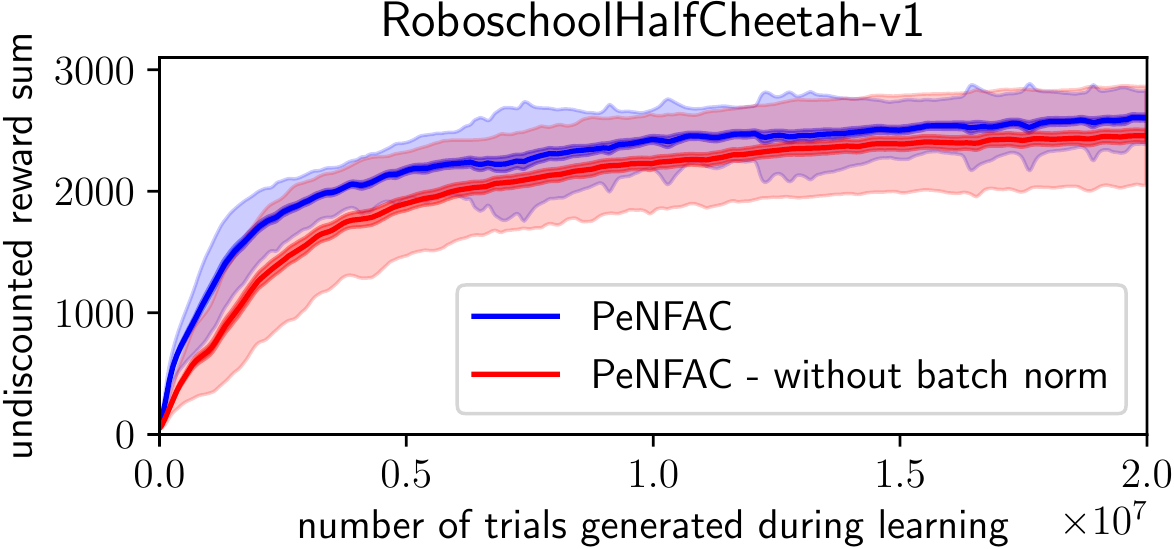}
    \end{center}
    \caption{Comparison of the different components ($\lambda$-returns, fitted value-iteration, CAC vs CACLA update, batch normalization) of the PeNFAC algorithm during the testing phase over the HalfCheetah environment and 60 seeds for each version. }
    \label{fig:penfaccomp}
\end{figure}

In Figures \ref{fig:penfacperf1}-\ref{fig:penfaccomp}, the lighter shade depicts one standard deviation around the average, while the darker shade is the standard deviation divided by the square root of the number of seeds. 
%For each algorithm, we start the learning process using 60 different seeds. 
%\pw{Each algorithm was run with 60 different seeds.}

In Figures \ref{fig:penfacperf1}-\ref{fig:penfacperf3}, PeNFAC outperforms DDPG and deterministic PPO during the testing phase. 
On Humanoid, even after optimizing the hyperparameters, we could not obtain the same results as those of \citeauthor{PPO} \shortcite{PPO}.
We conjecture that this may be explained as follows: 
1) the RoboschoolHumanoid moved from version 0 to 1, 
2) deterministic PPO 
\noindent might be less efficient than PPO, 
3) neither LinearAnneal for the exploration, nor adaptive Adam step size is present in the OpenAI Baseline implementation. 
However, we argue that the comparison should still be fair since PeNFAC also does not use those two components.
On Humanoid, we did not find a set of hyperparameters where DDPG could work correctly with both implementations.

\subsection{Components of PeNFAC}
\label{sec:ablationstudy}

In Figure~\ref{fig:penfaccomp}, we present an ablation analysis in the HalfCheetah domain to understand which components of the PenFAC algorithm are the most essential to its good performance.
From top to bottom plots of Figure~\ref{fig:penfaccomp}, we ran PenFAC with or without trust region, with or without $\lambda$-returns, with or without fitted value iteration, with CACLA update or CAC update, and finally with or without batch normalization.

It appears that $\lambda$-returns and fitted value iteration are the most needed, while the effect of batch normalization is small and mostly helps in the beginning of the learning. %may be questionable.

We also tried updating the actor every timestep without taking into account the sign of the advantage function (i.e., using SPG instead of CAC), but the algorithm was not able to learn at all.
This also demonstrates that the CAC update is an essential component of PenFAC.

%% file: tex/appendix.tex
%\section{Proofs} 

\onecolumn %can we?

\section{Proofs}

\subsection{Relation between DPG and CAC update for a given state}
\label{appendix:prooflimCAC}

For simplification, the proof of a single dimension $k$ of the parameter space is provided. To denote the $k$\textsuperscript{th} dimension of a vector $x$, we write $x_k$. If $x$ is a matrix, $x_{:,k}$ represents the $k$\textsuperscript{th} column vector. 
We will use the following result from \citeauthor{Silver2014} \shortcite{Silver2014}: 
\begin{flalign*} %\label{eq:stochdetertheo}
\lim_{\sigma \rightarrow 0} \nabla_\theta J(\pi_{\theta,\sigma}) = \nabla_\theta J(\mu_\theta).
\end{flalign*}
Thus, the following standard regularity conditions are required: $T, T_0, R, \mu, \pi, \nabla_a T, \nabla_a R, \nabla_\theta \mu$ are  continuous in all variables and bounded.
From this result, we derive the following equation for a fixed state $s$:
\begin{flalign*}
\lim_{\sigma \rightarrow 0} \int_{\mathcal{A}} A^\pi(s,a) \nabla_\theta \pi_{\theta,\sigma}(a|s)  da  = \nabla_a A^\mu(s,a) \big|_{a=\mu_\theta(s)} \nabla_\theta \mu_\theta(s).
\end{flalign*}

\noindent We first study the special case of $\Delta_{\text{DPG}}(s, \mu_\theta)_k = 0$ and want to show that ${\lim_{\sigma \rightarrow 0} \Delta_{\text{CAC}} (s, \mu_\theta)}_k$ is also zero:
%If $\nabla_\theta \mu_\theta(s) = 0$, it is trivial.
\begin{flalign*}
{\Delta_{\text{DPG}}(s, \mu_\theta)}_k = 0 \implies & \nabla_a A^\mu(s,a) \big|_{a=\mu_\theta(s)} {\nabla_\theta \mu_\theta(s)}_{:,k} = 0,\\
\implies & \lim_{\sigma \rightarrow 0} \int_{\mathcal{A}} A^\pi(s,a) {\nabla_\theta \pi_{\theta,\sigma}(a|s)}_{:,k} da = 0, \\
%\implies & \lim_{\sigma \rightarrow 0} \int \frac{1}{\sigma \sqrt{2 \pi}} e^{-\frac{1}{2} \frac{||b||_2^2}{\sigma^2}} \nabla_a A^\pi(s,a)\Big|_{a=\mu_\theta(s)+b} {\nabla_\theta \mu_\theta(s)}_{:,k} db = 0,\\
\implies & \lim_{\sigma \rightarrow 0} \frac{1}{\sigma^2} \int_{\mathcal{A}} \pi_{\theta, \sigma}(a|s) A^\pi(s,a) \big(a - \mu_\theta(s) \big)  {\nabla_\theta \mu_\theta(s)}_{:,k} da = 0,\\
%\implies & \nabla_a A^\pi(s,a)\Big|_{a=\mu_\theta(s)} \nabla_\theta \mu_\theta(s) = 0,\\
%\text{because $H\big(A^\pi(s,a)\big)$ is only a scalar value}\\
\implies & \lim_{\sigma \rightarrow 0} \frac{1}{\sigma^2} \int_{\mathcal{A}} \pi_{\theta, \sigma}(a|s) H\big(A^\pi(s,a)\big) A^\pi(s,a) \big(a - \mu_\theta(s) \big)  {\nabla_\theta \mu_\theta(s)}_{:,k} da = 0,\\
\implies & \lim_{\sigma \rightarrow 0} {\Delta_{\text{CAC}} (s, \mu_\theta)}_k = 0.
\end{flalign*}

\noindent Now, we study the more general case ${\Delta_{\text{DPG}}(s, \mu_\theta)}_k \neq 0$:
\begin{flalign*}
g_k^+(s, \mu_\theta) =& \frac{\lim_{\sigma \rightarrow 0} \Delta_{\text{CAC}} (s, \mu_\theta)_k}{\Delta_{\text{DPG}}(s, \mu_\theta)_k}, \\
=& \frac{\lim_{\sigma \rightarrow 0} \int_{\mathcal{A}} A^\pi(s,a) H(A^\pi(s,a)) \nabla_\theta {\pi_{\theta,\sigma}(a|s)}_{:,k} da}{ \lim_{\sigma \rightarrow 0} \int_{\mathcal{A}} A^\pi(s,a) \nabla_\theta {\pi_{\theta,\sigma}(a|s)}_{:,k} da }, \\
= & \lim_{\sigma \rightarrow 0} \frac{\int_{\mathcal{A}} A^\pi(s,a) H(A^\pi(s,a)) \nabla_\theta {\pi_{\theta,\sigma}(a|s)}_{:,k} da}{ \int_{\mathcal{A}} A^\pi(s,a) {\nabla_\theta \pi_{\theta,\sigma}(a|s)}_{:,k} da },\\
& \implies 0 \leq g_k^+(s, \mu_\theta) \leq 1.
\end{flalign*}

\subsection{Performance of a deterministic policy expressed from a Gaussian stochastic policy}
\label{appendix:fosl}
The proof is very similar to \cite{kakade2002approximately,Schulman2015} and easily extends to mixtures of stochastic and deterministic policies:
\begin{flalign*}
 & \int_{\mathcal{S}} d_\gamma^{\pi}(s) \int_\mathcal{A} \pi(a|s) A^\mu(s, a) da ds + \int_{\mathcal{S}} d_\gamma^{\tilde{\mu}}(s) A^\pi(s, \tilde{\mu}(s)) ds  = \\
& \int_{\mathcal{S}} d_\gamma^{\pi}(s) \int_\mathcal{A} \pi(a|s) \Big( R(s,a) + \gamma \mathbb{E}\big[V^\mu(s') | a\big] - V^\mu(s) \Big) da + 
    \int_{\mathcal{S}} d_\gamma^{\tilde{\mu}}(s) \Big( R(s,\tilde\mu(s)) + \gamma \mathbb{E}\big[V^\pi(s') | \tilde{\mu}(s) \big] - V^\pi(s) \Big) ds = \\
& J(\pi) + J(\tilde{\mu}) + \int_{\mathcal{S}} d_\gamma^{\pi}(s) \int_\mathcal{A} \pi(a|s)  \Big( \gamma \mathbb{E}\big[V^\mu(s') | a\big] - V^\mu(s) \Big) da +  \int_{\mathcal{S}} d_\gamma^{\tilde{\mu}}(s) \Big( \gamma \mathbb{E}\big[V^\pi(s') | \tilde{\mu}(s) \big] - V^\pi(s) \Big) ds = \\
& J(\pi) + J(\tilde{\mu}) + \int_{\mathcal{S}} d_\gamma^{\pi}(s) \Big( - V^\mu(s) +  \gamma \int_\mathcal{A} \pi(a|s) \mathbb{E}\big[V^\mu(s') | a\big]da \Big) - J(\pi) = \\
& J(\tilde{\mu}) - J(\mu).
\end{flalign*}

\subsection{Trust region for continuous deterministic policies}
\label{appendix:prooftrust}

For this theorem we also use the following standard regularity conditions: 
$I(\mathcal{S}) = \int_\mathcal{S} ds < \infty$ and $\Big|\Big| \tilde{\mu}(s) - \mu(s))\Big|\Big|_{2,\infty} < \infty$. $m$ denotes  the number of dimension of the action space.
We start from the two terms we want to bound:
\begin{flalign}
    &\Big| \int_{\mathcal{S}} d^{\tilde{\mu}}_\gamma(s) A^\pi(s, \tilde{\mu}(s)) - \int_{\mathcal{S}} d^{\pi}_\gamma(s) A^\pi(s, \tilde{\mu}(s))  \Big| = \notag \\
    & \Big| \int_{\mathcal{S}} \big( d^{\tilde{\mu}}_\gamma(s) - d^{\pi}_\gamma(s) \big) A^\pi(s, \tilde{\mu}(s)) \Big| \leq \notag \\
    & \int_{\mathcal{S}} \Big| d^{\tilde{\mu}}_\gamma(s) - d^{\pi}_\gamma(s) \Big| . \Big| A^\pi(s, \tilde{\mu}(s)) \Big| \leq  \notag \\
    & \epsilon \int_{\mathcal{S}} \Big| d^{\tilde{\mu}}_\gamma(s) - d^{\pi}_\gamma(s) \Big|,  \label{eq:proofstep3}
%    & \epsilon I(\mathcal{S}) \ \underset{s \in \mathcal{S}}{\operatorname{max}} \Big| d^{\tilde{\mu}}(s) - d^{\pi}(s) \Big|, \label{eq:proofstep3}
\end{flalign}
where $\epsilon = \text{max}_{s,a} |A^\pi(s,a)| $.

So, we need to bound the difference between $d^{\tilde{\mu}}$ and $d^{\pi}$ for a given state $s$: 
\begin{flalign}
& \Big| d^{\tilde{\mu}}_\gamma(s) - d^{\pi}_\gamma(s) \Big| = \notag \\ 
& \Big| \int_{\mathcal{S}} T_0(s_0) \Big( \sum^\infty_{t=0} \gamma^{t} p(s|s_0,t,\tilde{\mu}) - \sum^\infty_{\pw{t=0}}   \gamma^{t} p(s|s_0,t,\pi) \Big) ds_0 \Big| = \notag \\ 
& \Big| \int_{\mathcal{S}} T_0(s_0) \sum^\infty_{\pw{t=0}} \gamma^{t} \Big( p(s|s_0,t,\tilde{\mu}) - p(s|s_0,t,\pi) \Big) ds_0 \Big| \leq \notag \\
& \int_{\mathcal{S}} \Big| T_0(s_0) \Big|  \sum^\infty_{\pw{t=0}} \gamma^{t} \Big| p(s|s_0,t,\tilde{\mu}) - p(s|s_0,t,\pi) \Big| \notag ds_0 \leq \\
& \int_{\mathcal{S}} \sum^\infty_{\pw{t=0}} \gamma^{t} \Big| p(s|s_0,t,\tilde{\mu}) - p(s|s_0,t,\pi) \Big| ds_0 \leq \notag \\
& \int_{\mathcal{S}} \sum^\infty_{\pw{t=0}} \gamma^{t} \underset{t'>0}{\operatorname{max}} \Big| p(s|s_0,t',\tilde{\mu}) - p(s|s_0,t',\pi) \Big| ds_0 = \notag \\
& \frac{1}{1-\gamma} \int_{\mathcal{S}} \underset{t>0}{\operatorname{max}} \Big| p(s|s_0,t,\tilde{\mu}) - p(s|s_0,t,\pi) \Big| ds_0. \label{eq:proofstep4} %\leq \\
%& \frac{1}{1-\gamma} I(\mathcal{S}) \underset{s_0 \in \mathcal{S}}{\operatorname{max\ }} \underset{t}{\operatorname{max\ }} \Big| p(s|s_0,t,\tilde{\mu}) - p(s|s_0,t,\pi) \Big|. \label{eq:proofstep4}
\end{flalign}

Finally, we have to bound the difference between $ p(s|s_0,t,\tilde{\mu})$ and $ p(s|s_0,t,\pi) $.
To do so, we define $\tau = \{s_1, ..., s_t=s\}$, and $\mathcal{D}_\tau$ all the possible path from the state $s_1$ to the state $s_t=s$. 
%     & \pushright{ \int_\mathcal{A} \pi(a|s_{k-1}) T( s_k | s_{k-1}, a ) da \Big) d\tau \Big| \leq } \notag \\
\begin{flalign}
& \Big| p(s|s_0,t,\tilde{\mu}) - p(s|s_0,t,\pi)  \Big| = \notag \\
& \Big| \int_{\mathcal{D}_\tau} \prod_{k=1}^t \Big( T(s_k | s_{k-1}, \tilde{\mu}(s_{k-1})) \notag - \int_\mathcal{A} \pi(a|s_{k-1}) T( s_k | s_{k-1}, a ) da \Big) d\tau \Big| \leq \notag \\
& \int_{\mathcal{D}_\tau} \prod_{k=1}^t  \Big| T(s_k | s_{k-1}, \tilde{\mu}(s_{k-1})) \notag - \int_\mathcal{A} \pi(a|s_{k-1}) T( s_k | s_{k-1}, a ) da \Big| d\tau = \notag \\
& \int_{\mathcal{D}_\tau} \prod_{k=1}^t  \Big| \int_\mathcal{A} \pi(a|s_{k-1}) \big( T(s_k | s_{k-1}, \tilde{\mu}(s_{k-1})) \notag - T( s_k | s_{k-1}, a ) \big) da \Big| d\tau \leq \notag \\
& \int_{\mathcal{D}_\tau} \prod_{k=1}^t  \int_\mathcal{A} \pi(a|s_{k-1}) \Big| T(s_k | s_{k-1}, \tilde{\mu}(s_{k-1})) \notag -  T( s_k | s_{k-1}, a ) \Big| da d\tau \leq \notag
\end{flalign}
\begin{flalign}
& L \int_{\mathcal{D}_\tau} \prod_{k=1}^t  \int_\mathcal{A} \pi(a|s_{k-1}) \Big|\Big| \tilde{\mu}(s_{k-1}) - a\Big|\Big|_2 da d\tau = \label{eq:proofstep1} \\
& L \int_{\mathcal{D}_\tau} \prod_{k=1}^t  \int \frac{1}{(\sigma \sqrt{2 \pi})^m} e^{-\frac{1}{2\sigma^2} ||b||_2^2} \Big|\Big| \tilde{\mu}(s_{k-1}) - \mu(s_{k-1}) + b\Big|\Big|_2 db d\tau \leq  \label{eq:proofstep2} \\
& L \int_{\mathcal{D}_\tau} \prod_{k=1}^t  \int \frac{1}{(\sigma \sqrt{2 \pi})^m} e^{-\frac{1}{2\sigma^2} ||b||_2^2} \Big( \Big|\Big| \tilde{\mu}(s_{k-1}) - \mu(s_{k-1})\Big|\Big|_2 + \Big|\Big| b\Big|\Big|_2 \Big) db d\tau \leq \notag \\
& L \int_{\mathcal{D}_\tau} \prod_{k=1}^t \Big( \Big|\Big| \tilde{\mu}(s_{k-1}) - \mu(s_{k-1})\Big|\Big|_2 + \int \frac{1}{(\sigma \sqrt{2 \pi})^m} e^{-\frac{1}{2\sigma^2} ||b||_2^2} \Big|\Big| b\Big|\Big|_1 \Big) db d\tau = \notag \\
& L \int_{\mathcal{D}_\tau} \prod_{k=1}^t \Big( \Big|\Big| \tilde{\mu}(s_{k-1}) - \mu(s_{k-1})\Big|\Big|_2 + \frac{2m\sigma}{\sqrt{2 \pi}} \Big) d\tau \leq \notag \\
& L \int_{\mathcal{D}_\tau} \Big( \underset{s_k \in \tau}{\operatorname{max\ }} \Big|\Big| \tilde{\mu}(s_{k}) - \mu(s_{k})\Big|\Big|_2 \notag + \frac{2m\sigma}{\sqrt{2 \pi}} \Big)^t d\tau \leq \\
& I(\mathcal{S})^t L \Big( \underset{s_k \in \mathcal{S}}{\operatorname{max\ }} \Big|\Big| \tilde{\mu}(s_{k}) - \mu(s_{k})\Big|\Big|_2 + \frac{2m\sigma}{\sqrt{2 \pi}} \Big)^t. \label{eq:proofstep5}
\end{flalign}
To obtain (\ref{eq:proofstep1}), we use the assumption that the transition function is L-Lipschitz continuous with respect to the action and the L2 norm.
To obtain (\ref{eq:proofstep2}), we use (\ref{eq:hypo_polstoch}).
Equation \ref{eq:proofstep5} does no longer depend on $s$ and $s_0$, thus added to (\ref{eq:proofstep4}) and (\ref{eq:proofstep3}) it gives:
\begin{flalign}
 & \frac{\epsilon L}{1-\gamma} \underset{t>0}{\operatorname{max\ }} I(\mathcal{S})^{t+2} \Big( \Big|\Big| \tilde{\mu}(s) - \mu(s)\Big|\Big|_{2,\infty} + \frac{2m\sigma}{\sqrt{2 \pi}} \Big)^t \leq \notag  \\
 &\frac{\epsilon L}{1-\gamma} \underset{t>0}{\operatorname{max\ }} \Big( \Big|\Big| \tilde{\mu}(s) - \mu(s)\Big|\Big|_{2,\infty} + \frac{2m\sigma}{\sqrt{2 \pi}} \Big)^t. \label{eq:proofstep6}
\end{flalign}
To obtain (\ref{eq:proofstep6}), we suppose that $I(\mathcal{S})$ is smaller than 1. We can make this assumption without losing in generality: it would only affect the magnitude of the Lipschitz constant.
Thus if $ \big|\big| \tilde{\mu}(s) - \mu(s)\big|\big|_{2,\infty} + \frac{2m\sigma}{\sqrt{2 \pi}} $ stays smaller than $1$, the optimal $t$ will be $1$, and (\ref{eq:proofstep6}) could be reduced to: $$ \frac{\epsilon L}{1-\gamma} \Big( \Big|\Big| \tilde{\mu}(s) - \mu(s)\Big|\Big|_{2,\infty} + \frac{2m\sigma}{\sqrt{2 \pi}} \Big). $$

\section{Additional experiments on CACLA's update}

In those two experiments, we want to highlight the good performance of CACLA compared to SPG and DPG without neural networks. The main argument to use DPG instead of SPG is its efficiency when the action dimensions become large. In the first experiment, we study if CACLA suffers from the same variance problem as SPG. The second experiment supports our claim that CACLA is more robust than SPG and DPG when the approximation made by the critic is less accurate.

\subsection{Sensitivity to action space dimensionality}
\label{appendix:sensitivitydima}

We used a setup similar to that of \citeauthor{Silver2014} \shortcite{Silver2014}: those environments contain only one state and the horizon is fixed to one. They are designed such that the dimensionality of the action space can easily be controlled but there is only little bias in the critic approximation. The policy parameters are directly representing the action: $\mu_\theta(\cdot) = \theta$.

\begin{figure}[H]
    \begin{center}
        \includegraphics[width=0.7\textwidth]{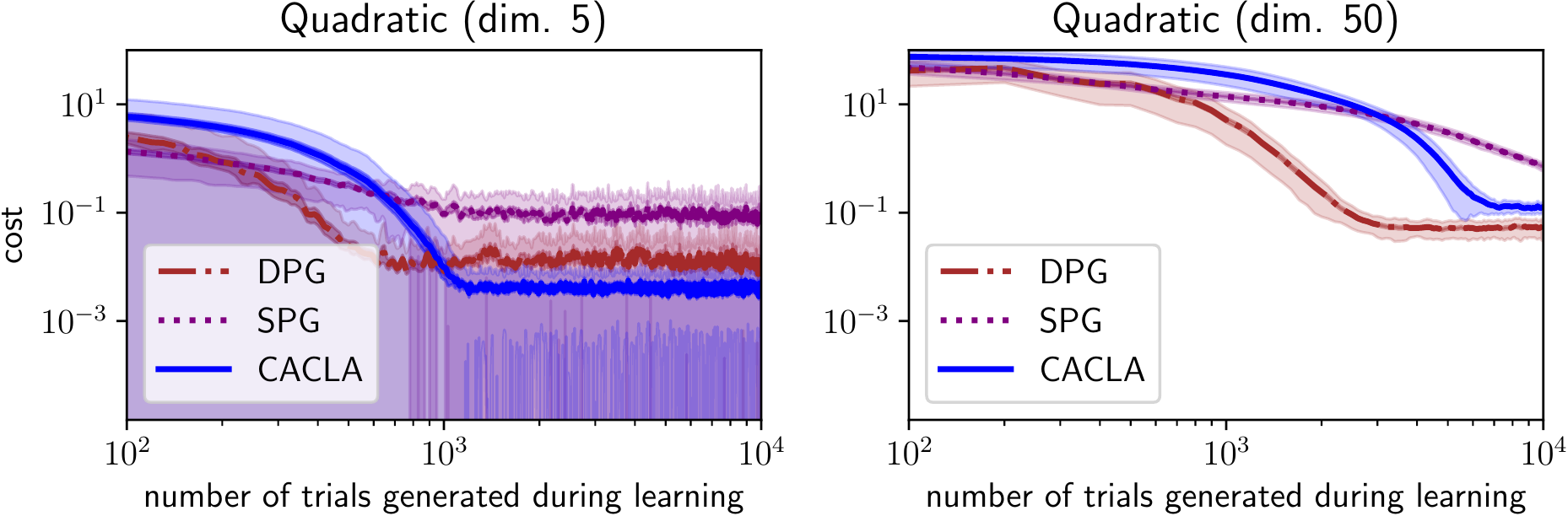}
        \includegraphics[width=0.7\textwidth]{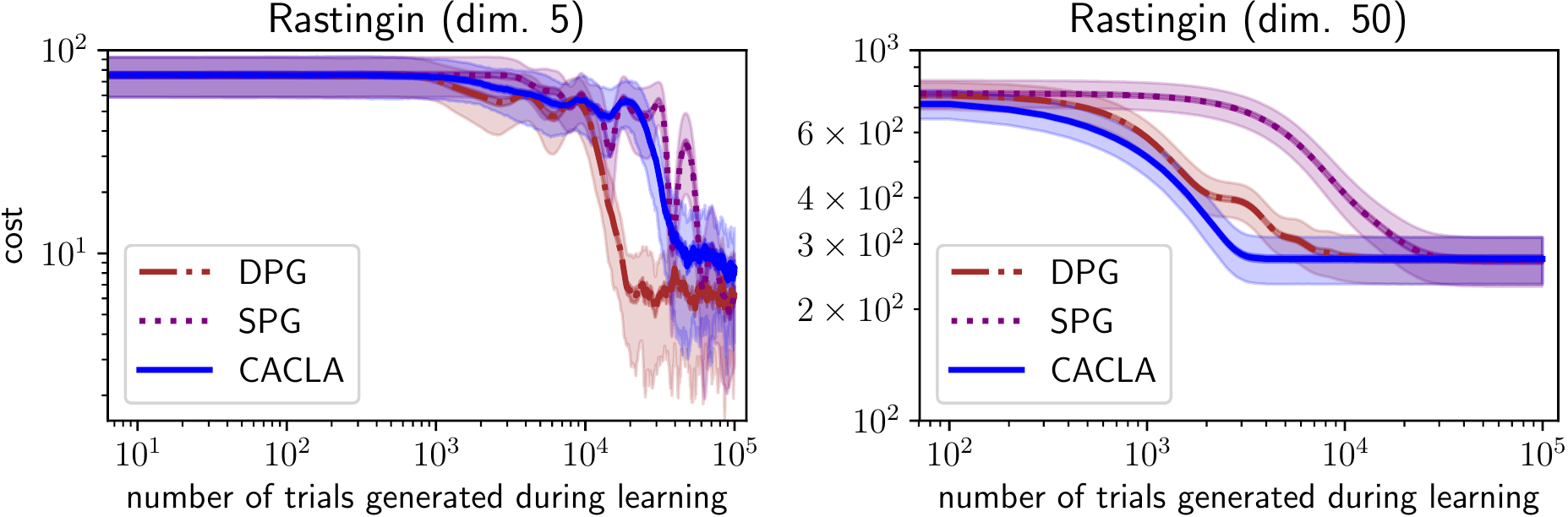}
        \includegraphics[width=0.7\textwidth]{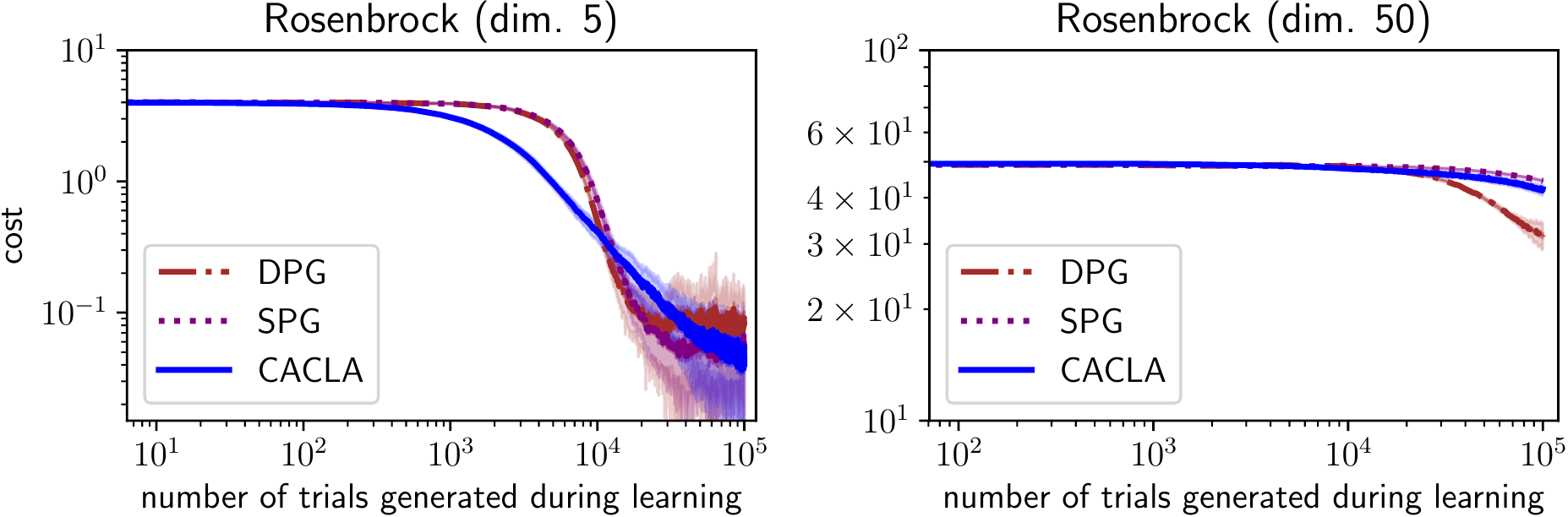}
    \end{center}
    \caption{Comparison of DPG, SPG and CACLA over three domains with 100 seeds for each algorithm. On the left, the action dimensions is 5 and 50 on the right.}
    \label{fig:1s}
\end{figure}

\noindent Compatible features are used to learn the Q value function for both SPG and DPG. For CACLA, the value function V is approximated through a single parameter. 
The Gaussian exploration noise and the learning rate of both the critic and actor have been optimized for each algorithm on each environment.
In Figure~\ref{fig:1s}, similarly to \citeauthor{Silver2014} \shortcite{Silver2014}, we observe that SPG is indeed more sensitive to larger action dimensions.
CACLA is also sensitive to this increase in dimensionality but not as much as SPG.
Finally, we also note that even if the solution of CACLA and DPG are not exactly the same theoretically, they are very similar in practice.

\subsection{Robustness to the critic approximation errors}

Compared to the previous experience, we introduce a bigger bias in the approximation of the critic by changing the application domains: the horizon is deeper and there is an infinite number of states.
The policy is represented as $\mu_\theta(s)=\phi(s) \cdot \theta$ where $\phi(s)$ are tiles coding features.

\begin{figure}[H]
    \begin{center}
        \includegraphics[width=0.33\linewidth]{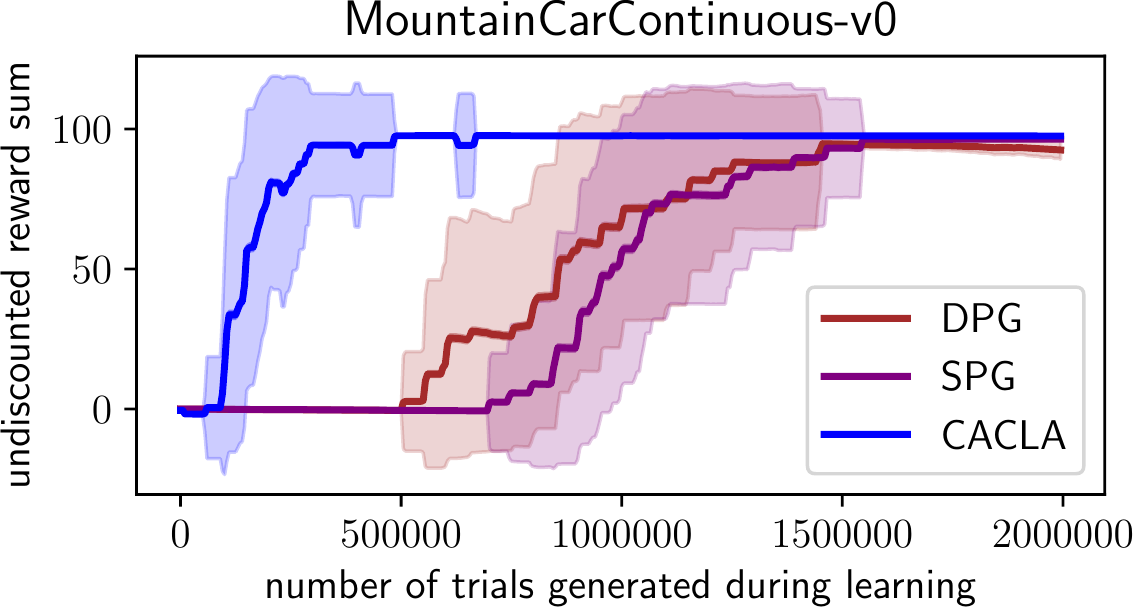}
        \includegraphics[width=0.33\linewidth]{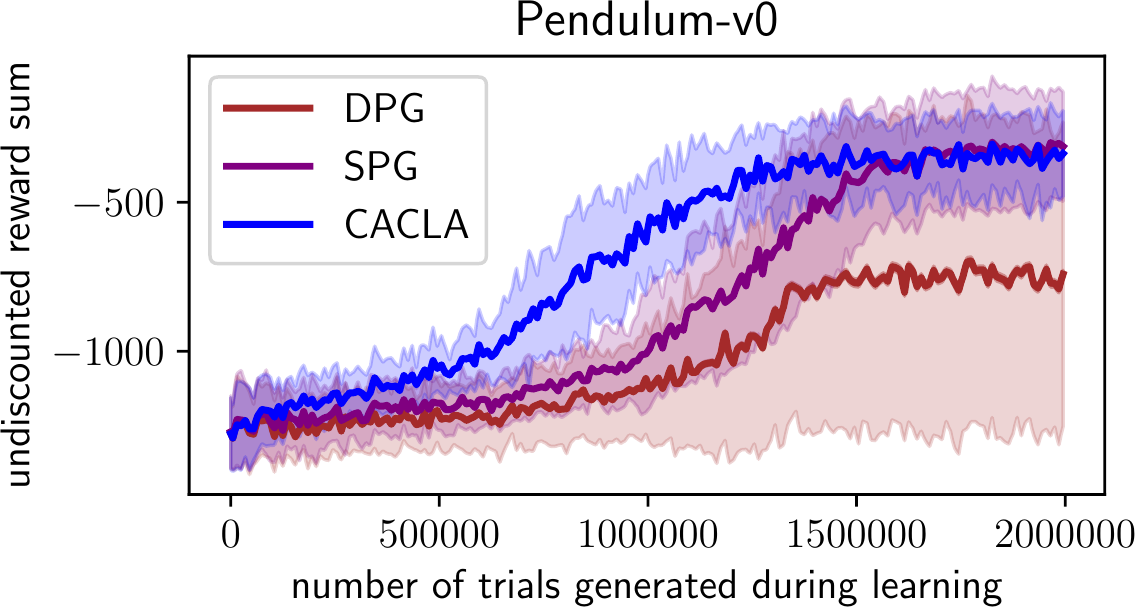}
        \includegraphics[width=0.33\linewidth]{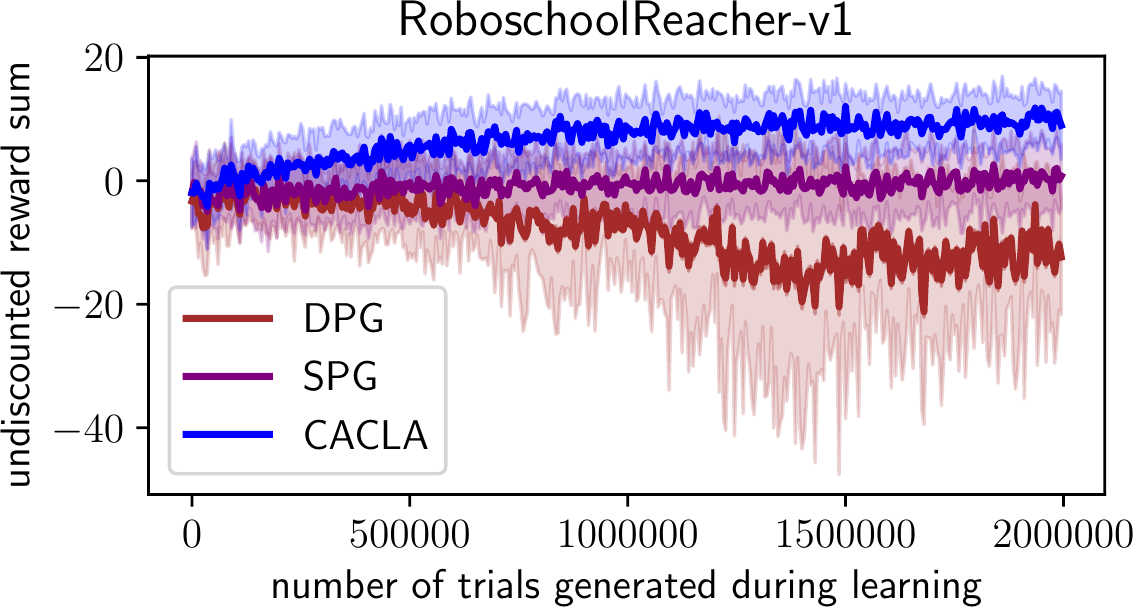}
    \end{center}
    \caption{Comparison of CACLA, DPG and SPG over two environments of OpenAI Gym and one environment of Roboschool (60 seeds are used for each algorithm). }
    \label{fig:cocla}
\end{figure}

In Figure~\ref{fig:cocla}, we observe that as soon as value functions become harder to learn, CACLA performs better than both SPG and DPG.

\section{Broader comparison between PeNFAC and NFAC}
\label{appendix:penfacvsnfac}

\begin{figure}[H]
    \begin{center}
        \includegraphics[width=0.48\linewidth]{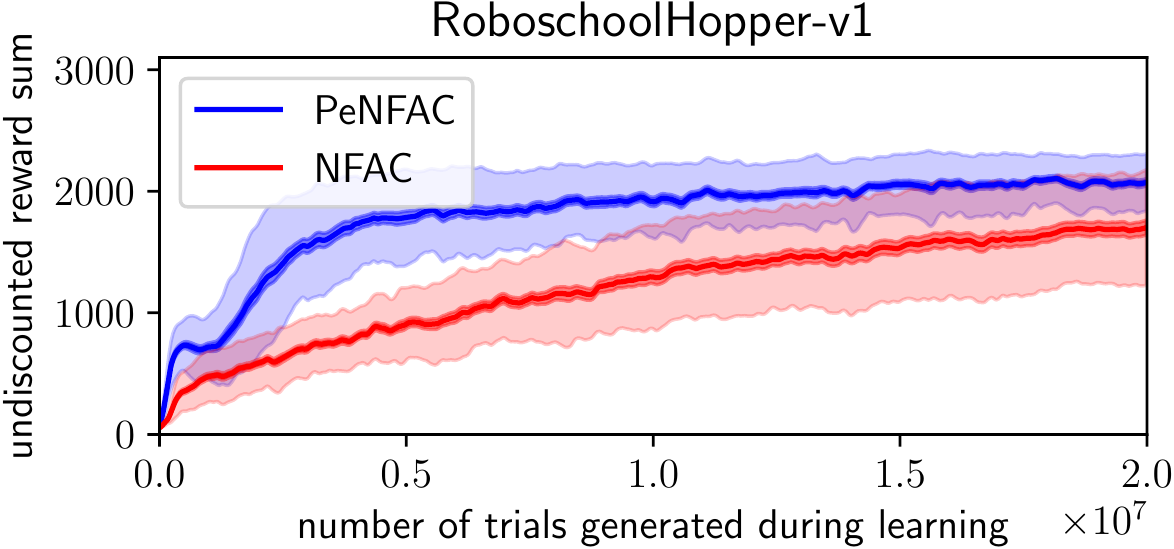}
        \includegraphics[width=0.48\linewidth]{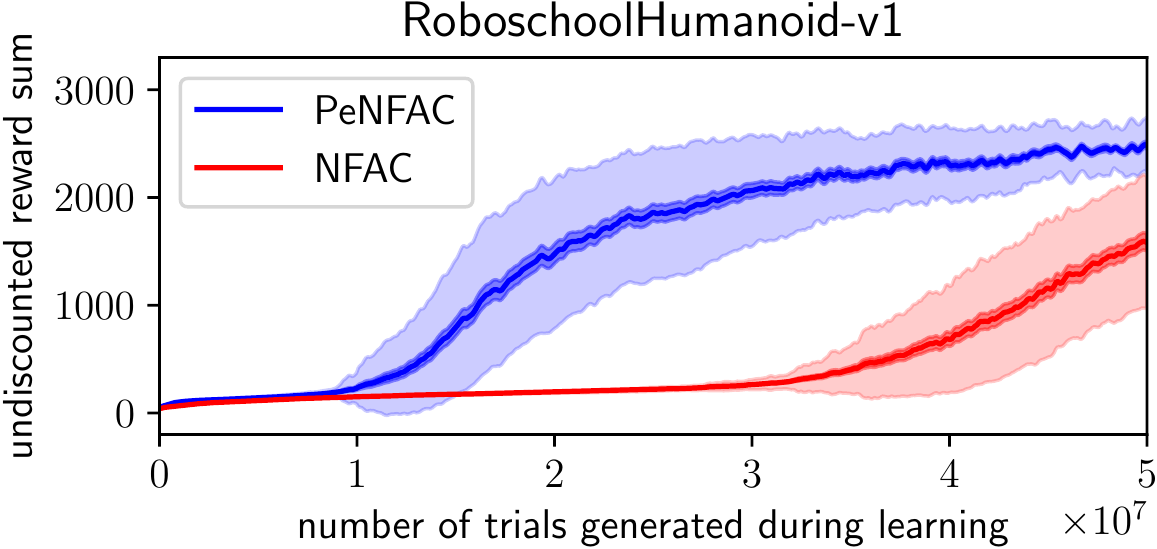}
    \end{center}
    \caption{Comparison of PeNFAC and NFAC over RoboschoolHopper and RoboschoolHumanoid with 60 seeds for each algorithm.}
    \label{fig:nfac}
\end{figure}

To avoid overloading previous curves, we did not report the performance of NFAC (except in the ablation study on the HalfCheetah environment). In Figure~\ref{fig:nfac}, we extend this study to two other domains of Roboschool: Hopper and Humanoid.

We observe that PeNFAC is significantly better than NFAC which demonstrates the efficiency of the trust region update combined with CAC.

\section{Impact of evaluating PPO with a deterministic policy}
\label{appendix:deterppo}

\begin{figure}[H]
    \begin{center}
        \includegraphics[width=0.48\linewidth]{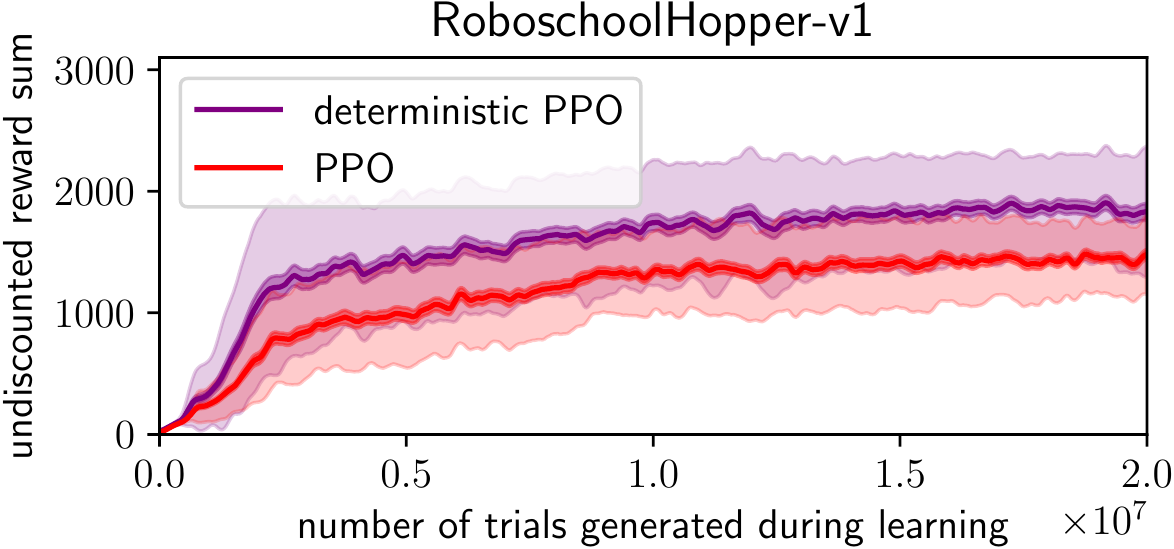}
        \includegraphics[width=0.48\linewidth]{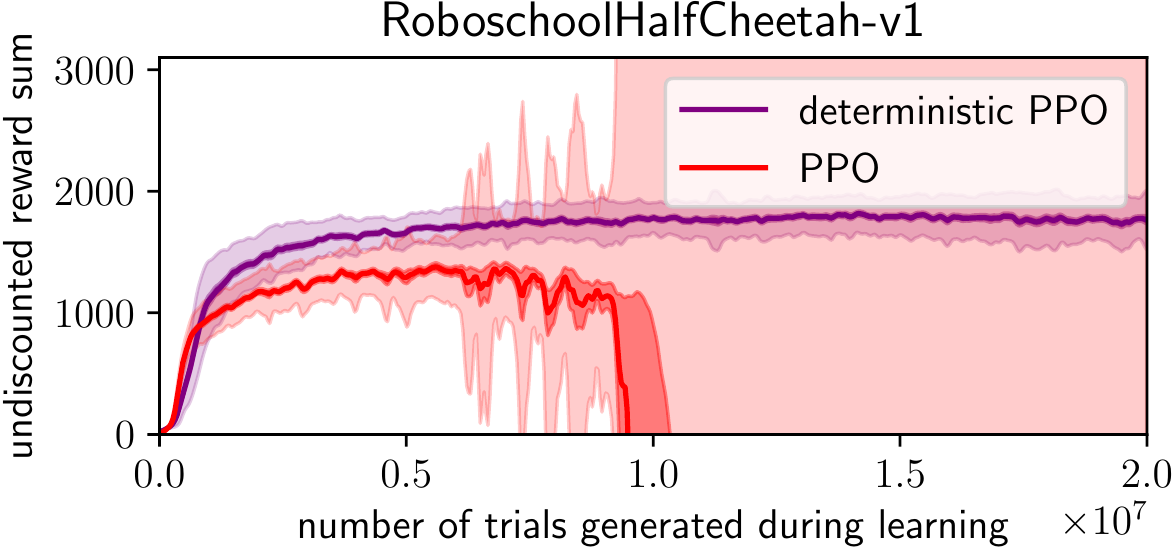}
    \end{center}
    \caption{Comparison of evaluating PPO with a deterministic policy instead of the stochastic policy produced by PPO.}
    \label{fig:deter_ppo}
\end{figure}

In Figure~\ref{fig:deter_ppo}, we observe that using a deterministic policy to evaluate the performance of PPO is not penalizing.
This is the only experiment of the paper where deterministic policies and stochastic policies are compared.

\section{Hyperparameters}
%add CPU time?
\label{appendix:hyperparam}
For the sake of reproducibility \cite{Henderson2017}, the hyperparameters used during the grid search are reported here. In Tables~\ref{tab:hyper1}-\ref{tab:hyper4}, "ho", "ha" and "hu" stand respectively for Hopper, HalfCheetah, and Humanoid Roboschool environments.

\begin{table}[H]
    \centering
    \begin{tabular}{c|c}
         $\gamma$ & $0.99$ \\
         Actor network & $64 \times 64$ \\
         Critic network & $64 \times 64$ \\
         Actor output activation & TanH \\
    \end{tabular}
    \caption{Set of hyperparameters used during the training with every algorithm.}
    \label{tab:hyperX}
\end{table}

\begin{table}[H]
    \centering
    \def\arraystretch{1.5}
    \begin{tabular}{c|c}
         Network hidden activation & $\underset{\text{ho,ha,hu}}{\text{Leaky ReLU} (0.01)}$, TanH \\
         Actor learning rate & $\underset{\text{ho,ha,hu}}{10^{-4}}$, $10^{-3}$, $10^{-2}$ \\
         Critic learning rate & $10^{-4}$, $\underset{\text{ho,ha,hu}}{10^{-3}}$, $10^{-2}$ \\
         Batch norm & first layer of the actor \\
         $d_{\text{target}}$ & $\underset{\text{ho,ha,hu}}{0.03}$, $0.01$, $0.005$ \\
         ADAM & $\underset{\text{ho,ha,hu}}{(0, 0.999, 10^{-8})}$, $(0.9, 0.999, 10^{-8})$ \\
         Number of ADAM iteration (actor) & 10, $\underset{\text{ho,ha,hu}}{30}$, $50$\\
         Number of ADAM iteration (critic) & 1\\
         $\lambda$ & $0$, $0.5$, $0.6$, $\underset{\text{ho,ha,hu}}{0.9}$, $0.95$, $0.97$ \\
         $\sigma^2$ (Truncated Gaussian law) & $0.01$, $0.05$, $0.1$, $\underset{\text{ho,ha,hu}}{0.2}$, $0.5$ \\
         Number fitted iteration & $1$, $\underset{\text{ho,ha,hu}}{10}$, $20$, $50$ \\
         Update each $x$ episodes & $1$, $2$, $3$, $\underset{\text{ha}}{5}$, $10$, $\underset{\text{ho}}{15}$, $20$, $30$, $\underset{\text{hu}}{50}$, $100$
    \end{tabular}
    \caption{Set of hyperparameters used during the training with PeNFAC.}
    \label{tab:hyper1}
\end{table}

\begin{table}[H]
    \centering
    \def\arraystretch{1.5}
    \begin{tabular}{c|c}
        Network hidden activation & $\underset{\text{ho,ha,hu}}{\text{TanH}}$, ReLu, Leaky ReLU (0.01) \\
        Layer norm & no \\
        ADAM & $(0.9, 0.999, 10^{-5})$ \\
        Entropy coefficient & 0 \\
        Clip range & 0.2 \\
        $\lambda$ & $\underset{\text{ho,ha}}{0.97}, \underset{\text{hu}}{0.95}$ \\
        Learning rate & $\underset{\text{hu}}{10^{-4}}, \underset{\text{ho,ha}}{3e^{-4}}$ \\
        nminibatches & $\underset{\text{hu}}{4}$, $\underset{\text{ho,ha}}{32}$ \\
        noptepochs & $4$, $\underset{\text{ho,ha}}{10}$, $15$, $\underset{\text{hu}}{50}$ \\
        nsteps & $2^{11}$, $\underset{\text{ha}}{2^{12}}$, $\underset{\text{ho}}{2^{13}}$, $\underset{\text{hu}}{2^{14}}$, $2^{15}$, $2^{16}$ \\
        sample used to make the policy more deterministic & 15 \\
    \end{tabular}
    \caption{Set of hyperparameters used during the training with PPO.}
    \label{tab:hyper2}
\end{table}

\begin{table}[H]
    \centering
    \def\arraystretch{1.5}
    \begin{tabular}{c|c}
        Network hidden activation & $\text{Leaky ReLU} (0.01)$ \\
        Actor learning rate & $10^{-4}$ \\
        Critic learning rate & $10^{-3}$ \\
        Batch norm & first layer of the actor \\
        ADAM & $\underset{\text{ho,ha,hu}}{(0, 0.999, 10^{-8})}$, $(0.9, 0.999, 10^{-8})$ \\
        L2 regularization of the critic & $\underset{\text{ha,ho}}{0.01}$, $\underset{\text{hu}}{\text{without}}$ \\
        Exploration & Gaussian ($0.2$), $\underset{\text{ha,ho,hu}}{\text{Ornstein Uhlenbeck} (0.001, 0.15, 0.01)}$\\
        Mini batch size & $32$, $64$, $\underset{\text{hu,ha,ho}}{128}$ \\
        Reward scale & $0.1$, $1$, $\underset{\text{hu,ha,ho}}{10}$ \\
        Soft update of target networks & $\underset{\text{hu}}{0.001}$, $\underset{\text{ha,ho}}{0.01}$ \\
        Replay memory & $10^6$ \\
        N-step returns & $\underset{\text{ha}}{1}$, $\underset{\text{hu,ho}}{5}$
    \end{tabular}
    \caption{Set of hyperparameters used during the training with DDPG (DDRL implementation).}
    \label{tab:hyper3}
\end{table}

\begin{table}[H]
    \centering
    \def\arraystretch{1.5}
    \begin{tabular}{c|c}
        Network hidden activation & $\underset{\text{ho,ha,hu}}{\text{ReLu}}$, TanH \\
        Actor learning rate & $10^{-4}$ \\
        Critic learning rate & $10^{-3}$ \\
        Layer norm & no \\
        ADAM & $(0.9, 0.999, 10^{-5})$ \\
        L2 regularization of the critic & $0.01$ \\
        Exploration & Ornstein Uhlenbeck ($0.2$), $\underset{\text{ho,ha,hu}}{\text{Parameter Space (0.2)}}$\\
        Mini batch size & $128$ \\
        Reward scale & $1$, $\underset{\text{ho,ha,hu}}{10}$ \\
        Soft update of target networks & $\underset{\text{ho,hu}}{0.001}$, $\underset{\text{ha}}{0.01}$ \\
        Replay memory & $10^6$ \\
        nb\_rollout\_steps & $10$,$\underset{\text{ho,ha,hu}}{100}$ \\
        nb\_train\_steps & $1$,$10$,$\underset{\text{ho,ha,hu}}{50}$ \\
    \end{tabular}
    \caption{Set of hyperparameters used during the training with DDPG (OpenAI baselines implementation).}
    \label{tab:hyper4}
\end{table}